\documentclass[review]{elsarticle}

\usepackage{multirow}
\usepackage{longtable}
\usepackage{booktabs}
\usepackage{graphicx}
\usepackage{bm}
\usepackage[tight,footnotesize]{subfigure}
\usepackage{url}
\usepackage{amsmath}

\usepackage{amsthm}

\theoremstyle{definition}
\newtheorem{definition}{Definition}[section]

\biboptions{sort&compress}

\usepackage{lscape}

\usepackage{lineno,hyperref}
\modulolinenumbers[5]

\let\oldequation\equation
\let\oldendequation\endequation

\renewenvironment{equation}
  {\linenomathNonumbers\oldequation}
  {\oldendequation\endlinenomath}

\journal{Journal of \LaTeX\ Templates}

\biboptions{authoryear}

\begin{document}

\begin{frontmatter}

\title{Graph Neural Network for Traffic Forecasting: A Survey}

\author{Weiwei Jiang\corref{mycorrespondingauthor}}
\address{Department of Electronic Engineering, Tsinghua University, Beijing, 100084, China}
\author{Jiayun Luo\corref{NTU}}
\address{School of Computer Science and Engineering, Nanyang Technological University, 639798, Singapore}
\cortext[mycorrespondingauthor]{Corresponding author. E-mail address: jiangweiwei@mail.tsinghua.edu.cn}
\cortext[NTU]{E-mail address: luoj0028@e.ntu.edu.sg}

\begin{abstract}
Traffic forecasting is important for the success of intelligent transportation systems. Deep learning models, including convolution neural networks and recurrent neural networks, have been extensively applied in traffic forecasting problems to model spatial and temporal dependencies. In recent years, to model the graph structures in transportation systems as well as contextual information, graph neural networks have been introduced and have achieved state-of-the-art performance in a series of traffic forecasting problems. In this survey, we review the rapidly growing body of research using different graph neural networks, e.g. graph convolutional and graph attention networks, in various traffic forecasting problems, e.g. road traffic flow and speed forecasting, passenger flow forecasting in urban rail transit systems, and demand forecasting in ride-hailing platforms. We also present a comprehensive list of open data and source codes for each problem and identify future research directions. To the best of our knowledge, this paper is the first comprehensive survey that explores the application of graph neural networks for traffic forecasting problems. We have also created a public GitHub repository where the latest papers, open data, and source codes will be updated.
\end{abstract}


\end{frontmatter}

\section{Introduction}
\label{sec:introduction}
Transportation systems are among the most important infrastructure in modern cities, supporting the daily commuting and traveling of millions of people. With rapid urbanization and population growth, transportation systems have become more complex. Modern transportation systems encompass road vehicles, rail transport, and various shared travel modes that have emerged in recent years, including online ride-hailing, bike-sharing, and e-scooter sharing. Expanding cities face many transportation-related problems, including air pollution and traffic congestion. Early intervention based on traffic forecasting is seen as the key to improving the efficiency of a transportation system and to alleviate transportation-related problems. In the development and operation of smart cities and intelligent transportation systems (ITSs), traffic states are detected by sensors (e.g. loop detectors) installed on roads, subway and bus system transaction records, traffic surveillance videos, and even smartphone GPS (Global Positioning System) data collected in a crowd-sourced fashion. Traffic forecasting is typically based on consideration of historical traffic state data, together with the external factors which affect traffic states, e.g. weather and holidays.

Both short-term and long-term traffic forecasting problems for various transport modes are considered in the literature. This survey focuses on the data-driven approach, which involves forecasting based on historical data. The traffic forecasting problem is more challenging than other time series forecasting problems because it involves large data volumes with high dimensionality, as well as multiple dynamics including emergency situations, e.g. traffic accidents. The traffic state in a specific location has both spatial dependency, which may not be affected only by nearby areas, and temporal dependency, which may be seasonal. Traditional linear time series models, e.g. auto-regressive and integrated moving average (ARIMA) models, cannot handle such spatiotemporal forecasting problems. Machine learning (ML) and deep learning techniques have been introduced in this area to improve forecasting accuracy, for example, by modeling the whole city as a grid and applying a convolutional neural network (CNN) as demonstrated by~\cite{jiang2018geospatial}. However, the CNN-based approach is not optimal for traffic foresting problems that have a graph-based form, e.g. road networks.

In recent years, graph neural networks (GNNs) have become the frontier of deep learning research, showing state-of-the-art performance in various applications~\citep{wu2020comprehensive}. GNNs are ideally suited to traffic forecasting problems because of their ability to capture spatial dependency, which is represented using non-Euclidean graph structures. For example, a road network is naturally a graph, with road intersections as the nodes and road connections as the edges. With graphs as the input, several GNN-based models have demonstrated superior performance to previous approaches on tasks including road traffic flow and speed forecasting problems. These include, for example, the diffusion convolutional recurrent neural network (DCRNN)~\citep{li2018dcrnn_traffic} and Graph WaveNet~\citep{wu2019graph} models. The GNN-based approach has also been extended to other transportation modes, utilizing various graph formulations and models.

To the best of the authors' knowledge, this paper presents the first comprehensive literature survey of GNN-related approaches to traffic forecasting problems. While several relevant traffic forecasting surveys exist~\citep{shi2018machine, pavlyuk2019feature, yin2020comprehensive, luca2020deep, fan2020deep, boukerche2020machine, manibardo2020deep, ye2020build, lee2020short, xie2020urban, george2020traffic, haghighat2020applications, boukerche2020artificial, tedjopurnomo2020survey, varghese2020deep}, most of them are not GNN-focused with only one exception~\citep{ye2020build}. For this survey, we reviewed 212 papers published in the years 2018 to 2020. Additionally, because this is a very rapidly developing research field, we also included preprints that have not yet gone through the traditional peer review process (e.g., arXiv papers) to present the latest progress. Based on these studies, we identify the most frequently considered problems, graph formulations, and models. We also investigate and summarize publicly available useful resources, including datasets, software, and open-sourced code, for GNN-based traffic forecasting research and application. Lastly, we identify the challenges and future directions of applying GNNs to the traffic forecasting problem. 

Instead of giving a whole picture of traffic forecasting, our aim is to provide a comprehensive summary of GNN-based solutions. This paper is useful for both the new researchers in this field who want to catch up with the progress of applying GNNs and the experienced researchers who are not familiar with these latest graph-based solutions. In addition to this paper, we have created an open GitHub repository on this topic~\footnote{\url{https://github.com/jwwthu/GNN4Traffic}}, where relevant content will be updated continuously.

Our contributions are summarized as follows:

1) \textit{Comprehensive Review}: We present the most comprehensive review of graph-based solutions for traffic forecasting problems in the past three years (2018-2020).

2) \textit{Resource Collection}: We provide the latest comprehensive list of open datasets and code resources for replication and comparison of GNNs in future work.

3) \textit{Future Directions}: We discuss several challenges and potential future directions for researchers in this field, when using GNNs for traffic forecasting problems. 

The remainder of this paper is organized as follows. In Section~\ref{sec:related}, we compare our work with other relevant research surveys. In Section~\ref{sec:problems}, we categorize the traffic forecasting problems that are involved with GNN-based models. In Section~\ref{sec:graphs}, we summarize the graphs and GNNs used in the reviewed studies. In Section~\ref{sec:resources}, we outline the open resources. Finally, in Section~\ref{sec:challenges}, we point out challenges and future directions.

\section{Related Research Surveys}
\label{sec:related}
In this section, we introduce the most recent relevant research surveys (most of which were published in 2020). The differences between our study and these existing surveys are pointed out when appropriate. We start with the surveys addressing wider ITS topics, followed by those focusing on traffic prediction problems and GNN application in particular.

Besides traffic forecasting, machine learning and deep learning methods have been widely used in ITSs as discussed in~\cite{haghighat2020applications, fan2020deep, luca2020deep}. In~\cite{haghighat2020applications}, GNNs are only mentioned in the task of traffic characteristics prediction. Among the major milestones of deep-learning driven traffic prediction (summarized in Figure 2 of~\cite{fan2020deep}), the state-of-the-art models after 2019 are all based on GNNs, indicating that GNNs are indeed the frontier of deep learning-based traffic prediction research.

Roughly speaking, five different types of traffic prediction methods are identified and categorized in previous surveys~\citep{xie2020urban, george2020traffic}, namely, statistics-based methods, traditional machine learning methods, deep learning-based methods, reinforcement learning-based methods, and transfer learning-based methods. Some comparisons between different categories have been considered, e.g., statistics-based models have better model interpretability, whereas ML-based models are more flexible as discussed in~\cite{boukerche2020artificial}. Machine learning models for traffic prediction are further categorized in~\cite{boukerche2020machine}, which include the regression model, example-based models (e.g., k-nearest neighbors), kernel-based models (e.g. support vector machine and radial basis function), neural network models, and hybrid models. Deep learning models are further categorized into five different generations in~\cite{lee2020short}, in which GCNs are classified as the fourth generation and other advanced techniques that have been considered but are not yet widely applied are merged into the fifth generation. These include transfer learning, meta learning, reinforcement learning, and the attention mechanism. Before these advanced techniques become mature in traffic prediction tasks, GNNs remain the state-of-the-art technique.

Some of the relevant surveys only focus on the progress of deep learning-based methods~\citep{tedjopurnomo2020survey}, while the others prefer to compare them with the statistics-based and machine learning methods~\citep{yin2020comprehensive, manibardo2020deep}. In~\cite{tedjopurnomo2020survey}, 37 deep neural networks for traffic prediction are reviewed, categorized, and discussed. The authors conclude that encoder-decoder long short term-memory (LSTM) combined with graph-based methods is the state-of-the-art prediction technique. A detailed explanation of various data types and popular deep neural network architectures is also provided, along with challenges and future directions for traffic prediction. Conversely, it is found that deep learning is not always the best modeling technique in practical applications, where linear models and machine learning techniques with less computational complexity can sometimes be preferable~\citep{manibardo2020deep}.

Additional research surveys consider aspects other than model selection. In~\cite{pavlyuk2019feature}, spatiotemporal feature selection and extraction pre-processing methods, which may also be embedded as internal model processes, are reviewed. A meta-analysis of prediction accuracy when applying deep learning methods to transport studies is given in~\cite{varghese2020deep}. In this study, apart from the models themselves, additional factors including sample size and prediction time horizon are shown to have a significant influence on prediction accuracy.

To the authors’ best knowledge, there are no existing surveys focusing on the application of GNNs for traffic forecasting. Graph-based deep learning architectures are reviewed in~\cite{ye2020build}, for a series of traffic applications, namely, traffic congestion, travel demand, transportation safety, traffic surveillance, and autonomous driving. Specific and practical guidance for constructing graphs in these applications is provided. The advantages and disadvantages of both GNNs and other deep learning models ,e.g. recurrent neural network (RNN), temporal convolutional network (TCN), Seq2Seq, and generative adversarial network (GAN), are examined. While the focus is not limited to traffic prediction problems, the graph construction process is universal in the traffic domain when GNNs are involved.

\section{Problems}
\label{sec:problems}
In this section, we discuss and categorize the different types of traffic forecasting problems considered in the literature. Problems are first categorized by the traffic state to be predicted. Traffic flow, speed, and demand problems are considered separately while the remaining types are grouped together under “other problems”. Then, the problem-types are further broken down into levels according to where the traffic states are defined. These include road-level, region-level, and station-level categories. 

Different problem types have different modelling requirements for representing spatial dependency. For the road-level problems, the traffic data are usually collected from sensors, which are associated with specific road segments, or GPS trajectory data, which are also mapped into the road network with map matching techniques. In this case, the road network topology can be seen as the graph to use, which may contain hundreds or thousands of road segments potentially. The spatial dependency may be described by the road network connectivity or spatial proximity. For the station-level problems, the metro or bus station topology can be taken as the graph to use, which may contain tens or hundreds of stations potentially. The spatial dependency may be described by the metro lines or bus routes. For the region-level problem, the regular or irregular regions are used as the nodes in a graph. The spatial dependency between different regions can be extracted from the land use purposes, e.g., from the points-of-interest data.

A full list of the traffic forecasting problems considered in the surveyed studies is shown in Table~\ref{tab:problems}. Instead of giving the whole picture of traffic forecasting research, only those problems with GNN-based solutions in the literature are listed in Table~\ref{tab:problems}.
\begin{center}
\tiny
\begin{longtable}{|p{2cm}|p{10cm}|}
\caption{Traffic forecasting problems in the surveyed studies.}
\label{tab:problems} \\
\hline
Problem & Relevant Studies \\
\hline
\endfirsthead

\multicolumn{2}{c}%
{{\bfseries \tablename\ \thetable{} -- continued from previous page}} \\
\hline Problem & Relevant Studies \\ \hline 
\endhead

\hline \multicolumn{2}{|r|}{{Continued on next page}} \\ \hline
\endfoot

\hline
\endlastfoot
    Road Traffic Flow & ~\cite{zhang2018kernel, wei2019dual, xu2020ge, guo2020optimized, zheng2020gman, pan2020spatio, pan2019urban, lu2019graph, mallick2020graph, zhang2020novel, zhang2020urban, bai2020adaptive, fang2019gstnet, huang2020lsgcn, wang2018efficient, zhang2020spatio, song2020spatial, xu2020spatial, wang2020traffic, chen2020gdcrn, lv2020temporal, kong2020stgat, fukuda2020short, zhang2020graphieee, boukerche2020performance, tang2020general, kang2019learning, guo2019attention, li2019hybrid, xu2019road, zhang2019graph, wu2018graph, sun2020traffic, wei2020spatial, li2020multi, cao2020spectral, bing2018spatio, yu2019st, li2020traffic, yin2020multi, chen2020tssrgcn, zhang2020attention, wang2020auto, xin2020internet, qu2020modeling, wang2020mtgcn, xie2020sast, huang2020short, guo2020dynamic, zhang2020spatial, fang2020meta, mengzhang2020spatial, tian2020st, xu2020traffic, chen2020gst} \\
    \hline
    Road OD Flow & ~\cite{xiong2020dynamic, ramadan2020traffic} \\
    \hline
    Intersection Traffic Throughput & ~\cite{sanchez2020gannster} \\
    \hline
    Regional Taxi Flow & ~\cite{zhou2020exploiting, sun2020predicting, chen2020dynamic, wang2018graph, peng2020spatial, zhou2019revisiting, wang2020multi, qiu2020topological} \\
    \hline
    Regional Bike Flow & ~\cite{zhou2020exploiting, sun2020predicting, wang2018graph, wang2020multi} \\
    \hline
    Regional Ride-hailing Flow & ~\cite{zhou2019revisiting} \\
    \hline
    Regional Dockless E-Scooter Flow & ~\cite{he2020dynamic} \\
    \hline
    Regional OD Taxi Flow & ~\cite{wang2020multi, yeghikyan2020learning} \\
    \hline
    Regional OD Bike Flow & ~\cite{wang2020multi} \\
    \hline
    Regional OD Ride-hailing Flow & ~\cite{shi2020predicting, wang2020urban, wang2019origin} \\
    \hline
    Station-level Subway Passenger Flow & ~\cite{fang2019gstnet, fang2020meta, peng2020spatial, ren2019transfer, li2018graph, zhao2020spatiotemporal, han2019predicting, zhang2020deep, zhang2020multi, li2020tensor, liu2020physical, ye2020multi, ou2020stp} \\
    \hline
    Station-level Bus Passenger Flow & ~\cite{fang2019gstnet, fang2020meta, peng2020spatial} \\
    \hline
    Station-level Shared Vehicle Flow & ~\cite{zhu2019multistep} \\
    \hline
    Station-level Bike Flow & ~\cite{he2020towards, chai2018bike} \\
    \hline
    Station-level Railway Passenger Flow & ~\cite{he2020gc} \\
    \hline
    Road Traffic Speed & ~\cite{li2018dcrnn_traffic, wu2019graph, zhang2018kernel, wei2019dual, xu2020ge, guo2020optimized, zheng2020gman, pan2020spatio, pan2019urban, lu2019graph, mallick2020graph, zhang2020novel, lv2020temporal, li2020multi, yin2020multi, guo2020dynamic, mengzhang2020spatial, chen2020dynamic, zhao2020spatiotemporal, zhu2020a3t, tang2020dynamic, james2020citywide, shin2020incorporating, liu2020graphsage, zhang2018gaan, zhang2019multistep, yu2019real, xie2019sequential, zhang2019spatial, guo2019multi, diao2019dynamic, cirstea2019graph, lu2019leveraging, zhang2019link, james2019online, ge2019temporal, ge2019traffic, zhang2019hybrid, lee2019ddp, shleifer2019incrementally, yu2020forecasting, ge2020global, lu2020lstm, yang2020relational, zhao2019t, cui2019traffic, chen2019gated, zhang2019gcgan, yu20193d, lee2019graph, bogaerts2020graph, wang2020forecast, cui2020graph, cui2020learning, guo2020an, zhou2020reinforced, cai2020traffic, zhou2020variational, wu2020connecting, chen2020multi, opolka2019spatio, mallick2020transfer, oreshkin2020fc, jia2020dynamic, sun2020constructing, guo2020short, xie2020deep, zhang2020spatial1, zhu2020ast, feng2020dynamic, zhu2020novel, fu2020bayesian, zhang2020graph, xie2020istd, park2020st, agafonov2020traffic, chen2020graph, lu2020spatiotemporal, jepsen2019graph, jepsen2020relational, bing2020integrating, lewenfus2020joint, zhu2020kst, liao2018deep, maas2020uncertainty, li2020two, song2020graph, zhao2020attention, guopeng2020dynamic, kim2020urban} \\
    \hline
    Road Travel Time & ~\cite{guo2020optimized, hasanzadeh2019piecewise, fang2020constgat, shao2020estimation, shen2020ttpnet} \\
    \hline
    Traffic Congestion & ~\cite{dai2020hybrid, mohanty2018graph, mohanty2020region, qin2020graph, han2020congestion} \\
    \hline
    Time of Arrival & ~\cite{hong2020heteta} \\
    \hline
    Regional OD Taxi Speed & ~\cite{hu2018recurrent} \\
    \hline
    Ride-hailing Demand & ~\cite{pian2020spatial, jin2020deep, li2020short, jin2020urban, geng2019multi, lee2019demand, bai2019stg2seq, geng2019spatiotemporal, bai2019spatio, ke2020joint, li2020sdcn} \\
    \hline
    Taxi Demand & ~\cite{lee2019demand, bai2019stg2seq, bai2019spatio, ke2019predicting, hu2020stochastic, zheng2020spatial, xu2019incorporating, davis2020grids, chen2020multitask, du2020traffic, li2020forecaster, wu2020multi, ye2020coupled} \\
    \hline
    Shared Vehicle Demand & ~\cite{luo2020d3p} \\
    \hline
    Bike Demand & ~\cite{lee2019demand, bai2019stg2seq, bai2019spatio, du2020traffic, ye2020coupled, chen2020context, wang2020spatial, qin2020resgcn, xiao2020demand, yoshida2019practical, guo2019bikenet, kim2019graph, lin2018predicting} \\
    \hline
    Traffic Accident & ~\cite{zhou2020riskoracle, yu2020deep, zhang2020multi1, zhou2020foresee} \\
    \hline
    Traffic Anomaly & ~\cite{Liu2020stmfm} \\
    \hline
    Parking Availability & ~\cite{zhang2020semijournal, yang2019deep, zhang2020semi} \\
    \hline
    Transportation Resilience & ~\cite{wang2020evaluation} \\
    \hline
    Urban Vehicle Emission & ~\cite{xu2020spatiotemporal} \\
    \hline
    Railway Delay & ~\cite{heglund2020railway} \\
    \hline
    Lane Occupancy & ~\cite{wright2019neural} \\
    \hline
\end{longtable}
\end{center}

Generally speaking, traffic forecasting problems are challenging, not only for the complex temporal dependency, but only for the complex spatial dependency. While many solutions have been proposed for dealing with the time dependency, e.g., recurrent neural networks and temporal convolutional networks, the problem to capture and model the spatial dependency has not been fully solved. The spatial dependency, which refers to the complex and nonlinear relationship between the traffic state in one particular location with other locations. This location could be a road intersection, a subway station, or a city region. The spatial dependency may not be local, e.g., the traffic state may not only be affected by nearby areas, but also those which are far away in the spatial range but connected by a fast transportation tool. The graphs are necessary to capture such kind of spatial information as we would discuss in the next section.

Before the usage of graph theories and GNNs, the spatial information is usually extracted by multivariate time series models or CNNs. Within a multivariate time series model, e.g., vector autoregression, the traffic states collected in different locations or regions are combined together as multivariate time series. However, the multivariate time series models can only extract the linear relationship among different states, which is not enough for modeling the complex and nonlinear spatial dependency. CNNs take a step further by modeling the local spatial information, e.g., the whole spatial range is divided into regular grids as the two-dimensional image format and the convolution operation is performed in the neighbor grids. However, the CNN-based approach is bounded to the case of Euclidean structure data, which cannot model the topological structure of the subway network or the road network.

Graph neural networks bring new opportunities for solving traffic forecasting problems, because of their strong learning ability to capture the spatial information hidden in the non-Euclidean structure data, which are frequently seen in the traffic domain. Based on graph theories, both nodes and edges have their own attributes, which can be used further in the convolution or aggregation operations. These attributes describe different traffic states, e.g., volume, speed, lane numbers, road level, etc. For the dynamic spatial dependency, dynamic graphs can be learned from the data automatically. For the case of hierarchical traffic problems, the concepts of super-graphs and sub-graphs can be defined and further used.

\subsection{Traffic Flow}
Traffic flow is defined as the number of vehicles passing through a spatial unit, such as a road segment or traffic sensor point, in a given time period. An accurate traffic flow prediction is beneficial for a variety of applications, e.g., traffic congestion control, traffic light control, vehicular cloud, etc~\citep{boukerche2020machine}. For example, traffic light control can reduce vehicle staying time at the road intersections, optimizing the traffic flow, and reducing traffic congestion and vehicle emission.

We consider three levels of traffic flow problems in this survey, namely, road-level flow, region-level flow, and station-level flow. 

Road-level flow problems are concerned with traffic volumes on a road and include \textit{road traffic flow}, \textit{road origin-destination (OD) Flow}, and \textit{intersection traffic throughput}. In road traffic flow problems, the prediction target is the traffic volume that passes a road sensor or a specific location along the road within a certain time period (e.g. five minutes). In the road OD flow problem, the target is the volume between one location (the origin) and another (the destination) at a single point in time. The intersection traffic throughput problem considers the volume of traffic moving through an intersection.

Region-level flow problems consider traffic volume in a region. A city may be divided into regular regions (where the partitioning is grid-based) or irregular regions (e.g. road-based or zip-code-based partitions). These problems are classified by transport mode into \textit{regional taxi flow}, \textit{regional bike flow}, \textit{regional ride-hailing flow}, \textit{regional dockless e-scooter flow}, \textit{regional OD taxi flow}, \textit{regional OD bike flow}, and \textit{regional OD ride-hailing flow} problems.

Station-level flow problems relate to the traffic volume measured at a physical station, for example, a subway or bus station. These problems are divided by station type into \textit{station-level subway passenger flow}, \textit{station-level bus passenger flow}, \textit{station-level shared vehicle flow}, \textit{station-level bike flow}, and \textit{station-level railway passenger flow} problems.

Road-level traffic flow problems are further divided into cases of unidirectional and bidirectional traffic flow, whereas region-level and station-level traffic flow problems are further divided into the cases of inflow and outflow, based on different problem formulations.

While traffic sensors have been successfully used, data collection for traffic flow information is still a challenge when considering the high costs in deployment and maintenance of traffic sensors. Another potential approach is using the pervasive mobile and IoT devices, which have a lower cost generally, e.g., GPS sensors. However, challenges still exist when considering the data quality problems frequently seen in GPS data, e.g., missing data caused by unstable communication links.

The traffic light is another source of challenges for various traffic prediction tasks. Short-term traffic flow fluctuation and the spatial relation change between two road segments can be caused by the traffic light. The way of controlling the traffic light may be different in different time periods, causing an inconsistent traffic flow pattern.

\subsection{Traffic Speed}
Traffic speed is another important indicator of traffic state with potential applications in ITS systems, which is defined as the average speed of vehicles passing through a spatial unit in a given time period. The speed value on the urban road can reflect the crowdedness level of road traffic. For example, Google Maps visualizes this crowdedness level from crowd-sourcing data collected from individual mobile devices and in-vehicle sensors. A better traffic speed prediction is also useful for route navigation and estimation-of-arrival applications.

We consider two levels of traffic speed problems in this survey, namely, road-level and region-level problems. We also include travel time and congestion predictions in this category because they are closely correlated to traffic speed. Travel time prediction is useful for passengers to plan their commuting time and for drivers to select fast routes, respectively. Traffic congestion is one of the most important and urgent transportation problems in cities, which brings significant time loss, air pollution and energy waste. The congestion prediction results can be used to control the road conditions and optimize vehicle flow, e.g., with traffic signal control. In several studies, traffic congestion is judged by a threshold-based speed inference. The specific road-level speed problem categories considered are \textit{road traffic speed}, \textit{road travel time}, \textit{traffic congestion}, and \textit{time of arrival} problems; while the region-level speed problem considered is \textit{regional OD taxi speed}.

Traffic speed is concerned in both urban roads and freeways. However, the challenges differ in these two different scenarios. Freeways have a few traffic signals or on/off-ramps, making the prediction easier than the urban case. And the challenge mainly comes from the complex temporal dependency. More complex traffic networks exist in urban roads with more complicated connection patterns and abrupt changes. For example, different road segments may have different speed limit values and the allowed vehicle types. Besides the complex temporal dependency, modeling the spatial dependency becomes a bigger challenge for urban traffic speed forecasting.

\subsection{Traffic Demand}
Traffic demand prediction is a key component for taxi and ride-hailing services to be successful, which benefits these service providers to allocate limited available transportation resources to those urban areas with a higher demand. For passengers, traffic demand prediction encourages the consideration of various transportation forms, e.g., taking the public transit service when taxi or ride-hailing services are in short supply.

Traffic demand refers to the potential demand for travel, which may or may not be fulfilled completely. For example, on an online ride-hailing platform, the ride requests sent by passengers represent the demand, whereas only a subset of these requests may be served depending on the supply of drivers and vehicles, especially during rush hours. Accurate prediction of travel demand is a key element of vehicle scheduling systems (e.g. online ride-hailing or taxi dispatch platforms). However, in some cases, it is difficult to collect the potential travel demand from passengers and a compromise method using transaction records as an indication of the traffic demand is used. In such cases the real demand may be underestimated. Based on transport mode, the traffic demand problems considered include \textit{ride-hailing demand}, \textit{taxi demand}, \textit{shared vehicle demand}, and \textit{bike demand}.

\subsection{Other Problems}
In addition to the above three categories of traffic forecasting problems, GNNs are also being applied to the following problems.

\textit{Traffic accident} and \textit{Traffic anomaly}: the target is to predict the traffic accident number reported to the police system. Traffic anomaly is the major cause of traffic delay and a timely detection and prediction would help the administrators to identify the situation and turn the traffic situation back to normal as quickly as possible. A traffic accident is usually an accident in road traffic involving different vehicles, which may cause significant loss of life and property. The traffic anomaly has a broader definition that deviates from the normal traffic state, e.g., the traffic jam caused by a traffic accident or a public procession.

\textit{Parking availability}: the target is to predict the availability of vacant parking space for cars in the streets or in a car parking lot.

\textit{Urban vehicle emission}: while not directly related to traffic states, the prediction of urban vehicle emission is considered in~\cite{xu2020spatiotemporal}. Urban vehicle emission refers to the emission produced by motor vehicles, e.g., those use internal combustion engines. Urban vehicle emission is a major source of air pollutants and its amount is affected by different traffic states, e.g., the excess emission would be created in traffic congestion situations.
    
\textit{Railway delay}: the delay time of specific routes in the railway system is considered in ~\cite{heglund2020railway}.
    
\textit{Lane occupancy}: With simulated traffic data, lane occupancy has been measured and predicted~\citep{wright2019neural}.

\section{Graphs and Graph Neural Networks}
\label{sec:graphs}
In this section, we summarize the types of graphs and GNNs used in the surveyed studies, focusing on GNNs that are frequently used for traffic forecasting problems. The contributions of this section include an organized approach for classifying the different traffic graphs based on the domain knowledge, and a summary of the common ways for constructing adjacency matrices, which may not be encountered in other neural networks before and would be very helpful for those who would like to use graph neural networks. The different GNN structures already used for traffic forecasting problems are briefly introduced in this section too. For a wider and deeper discussion of GNNs, refer to~\cite{wu2020comprehensive, zhou2018graph, zhang2020deeplearning}.

\subsection{Traffic Graphs}
\label{sec:graphs}
\subsubsection{Graph Construction}
A graph is the basic structure used in GNNs. It is defined as $G=(V, E, A)$, where $V$ is the set of vertices or nodes, $E$ is the set of edges between the nodes, and $A$ is the adjacency matrix. Both nodes and edges can be associated with different attributes in different GNN problems. Element $a_{ij}$ of $A$ represents the “edge weight” between nodes $i$ and $j$. For a binary connection matrix $A$, $a_{ij} = 1$ if there is an edge between nodes $i$ and $j$ in $E$, and $a_{ij} = 0$ otherwise. If $A$ is symmetric, the corresponding graph $G$ is defined as undirected. Otherwise, $G$ is directed, when the edge only exists in one direction between a node pair.

For simplicity, we assume that the traffic state is associated with the nodes. The other case with edges can be derived similarly. In practice, the traffic state is collected or aggregated in discrete time steps, e.g. five minutes or one hour, depending on the specific scenario.

For a single time step $t$, we denote the node feature matrix as $\chi_t \in {R}^{N \times d}$, where $N$ is the number of nodes and $d$ is the dimension of the node features, i.e., the number of traffic state variables. Now we are ready to give a formal definition of traffic graph.

\begin{definition}[Traffic Graph]
A traffic graph (with node features) is defined as a specific type of graph $G=(V, E, A)$, where $V$ is the node set, $E$ is the edge set, and $A$ is the adjacency matrix. For a single time step $t$, the node feature matrix $\chi_t \in {R}^{N \times d}$ for $G$ contains specific traffic states, where $N$ is the number of nodes and $d$ is the number of traffic state variables.
\end{definition}

Then we give a formal definition of graph-based traffic forecasting problem without leveraging external factors firstly.

\begin{definition}[Graph-based Traffic Forecasting]
A graph-based traffic forecasting (without external factors) is defined as follows: find a function $f$ which generates $y=f(\mathbf{\chi};G)$, where $y$ is the traffic state to be predicted, $\mathbf{\chi}=\{\chi_1, \chi_2, ..., \chi_T\}$ is the historical traffic state defined on graph $G$, and $T$ is the number of time steps in the historical window size.
\end{definition}

\begin{figure}[!htb]
    \centering
    \includegraphics[width=\textwidth]{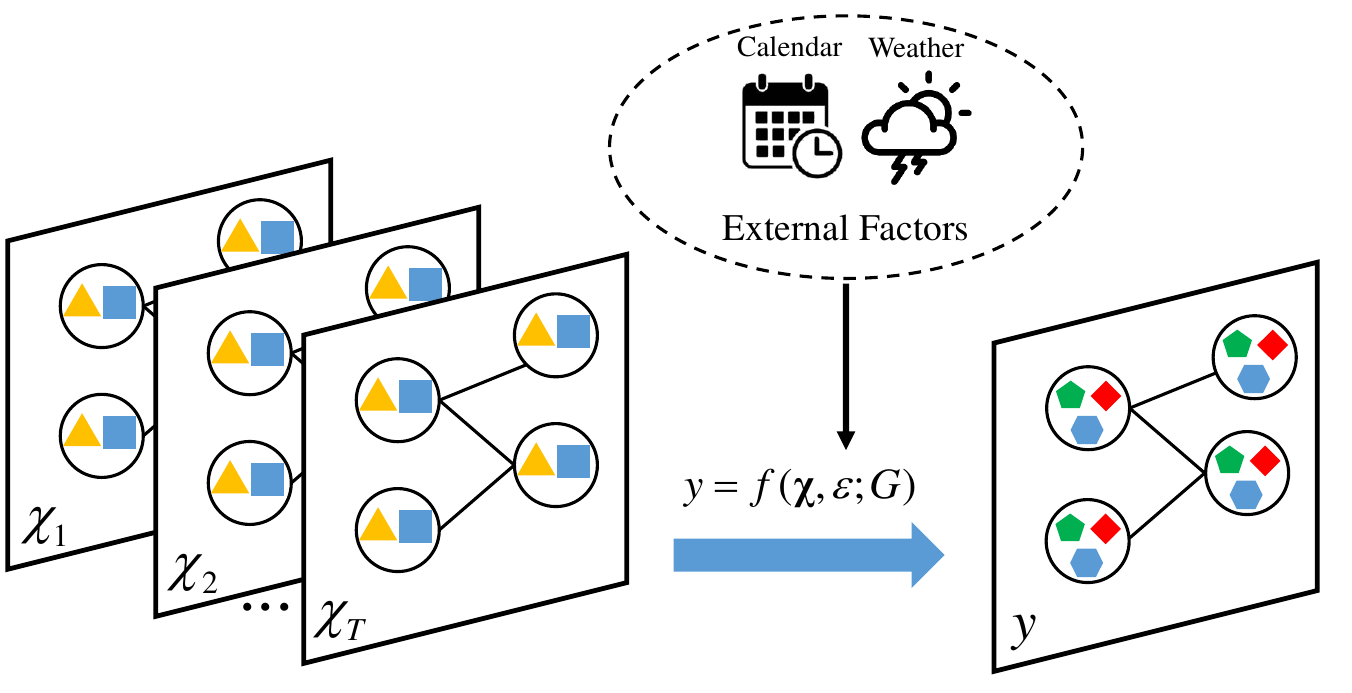}
    \caption{The single-step graph-based traffic forecasting problem. Adapted from~\cite{ye2020build} with external factors added.}
    \label{fig:fig1}
\end{figure}

In \textit{single step forecasting}, the traffic state in the next time step only is predicted, whereas in \textit{multiple step forecasting} the traffic state several time steps later is the prediction target. As mentioned in Section~\ref{sec:introduction}, traffic states can be highly affected by external factors, e.g. weather and holidays. The forecasting problem formulation, extended to incorporate these external factors, takes the form $y=f(\mathbf{\chi}, \varepsilon;G)$, where $\varepsilon$ represents the external factors. Figure~\ref{fig:fig1} demonstrates the graph-based traffic forecasting problem, where different color patches represent different traffic variables.

Various graph structures are used to model traffic forecasting problems depending on both the forecasting problem-type and the traffic datasets available. These graphs can be pre-defined static graphs, or dynamic graphs continuously learned from the data. The static graphs can be divided into two types, namely, natural graphs and similarity graphs. Natural graphs are based on a real-world transportation system, e.g. the road network or subway system; whereas similarity graphs are based solely on the similarity between different node attributes where nodes may be virtual stations or regions.

We categorize the existing traffic graphs into the same three levels used in Section~\ref{sec:problems}, namely, road-level, region-level and station-level graphs.

\begin{figure*}[!htb]
\centering
\subfigure[][]{
    \includegraphics[width=0.8\textwidth]{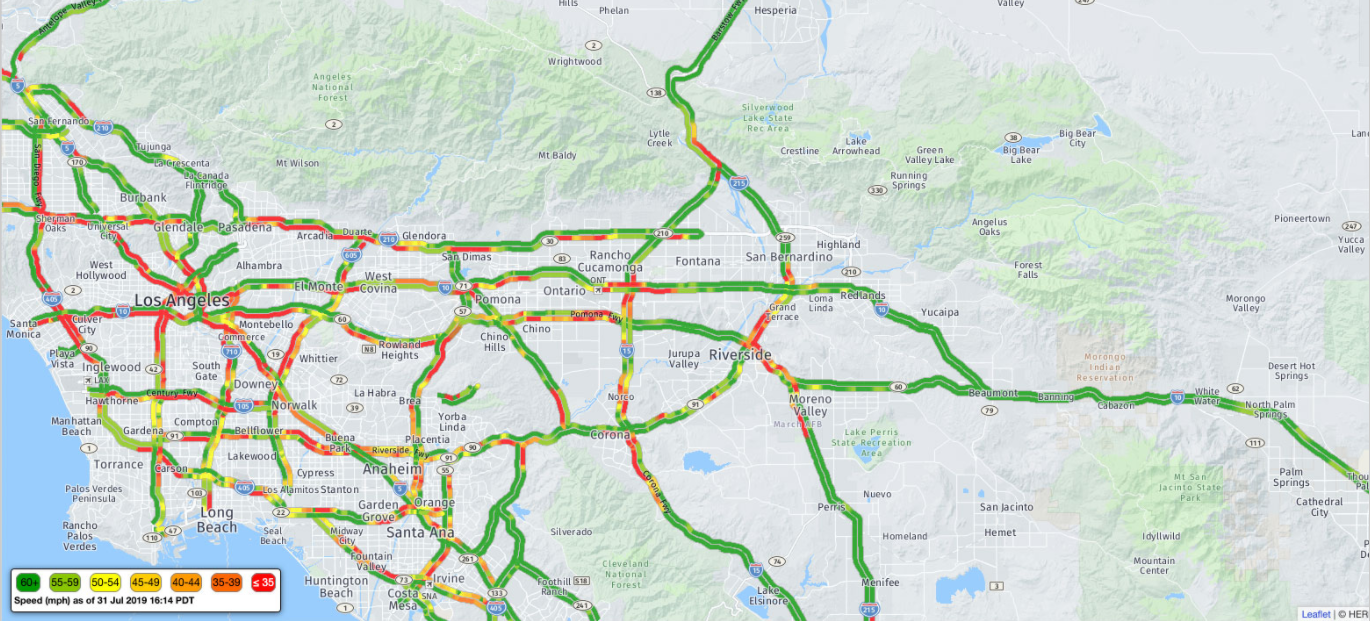}
    \label{fig:pems}
} \\
\subfigure[][]{
    \includegraphics[width=0.95\textwidth]{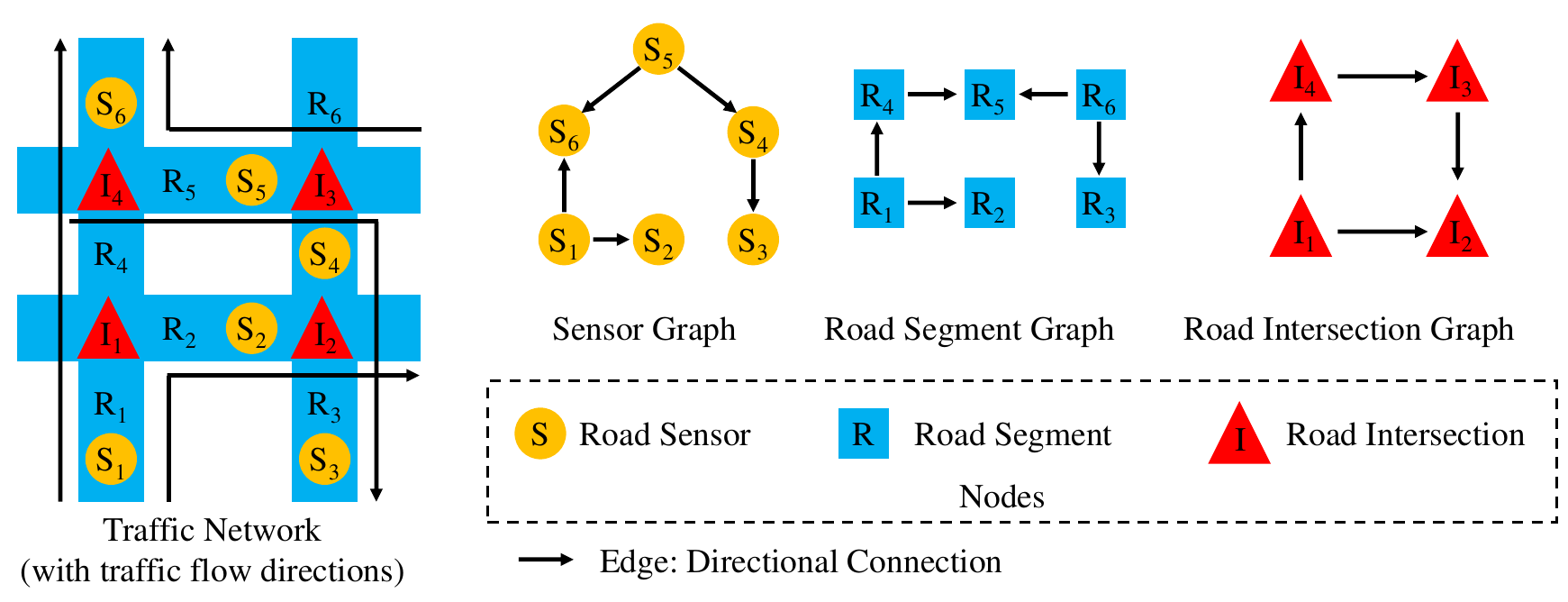}
    \label{fig:fig2a}
} 
\caption[]{The real-world case and example of road-level graphs.
    \subref{fig:pems} The road network in the Performance Measurement System (PeMS) where each sensor is a node. Source: \url{http://pems.dot.ca.gov/}.
    \subref{fig:fig2a} The road-level graph examples. Adapted from~\cite{ye2020build}.}
\label{fig:graphs-road}
\end{figure*}

\textit{Road-level graphs}. These include \textit{sensor graphs}, \textit{road segment graphs}, \textit{road intersection graphs}, and \textit{road lane graphs}. Sensor graphs are based on traffic sensor data (e.g. the PeMS dataset) where each sensor is a node, and the edges are road connections. The other three graphs are based on road networks with the nodes formed by road segments, road intersections, and road lanes, respectively. The real-world case and example of road-level graphs are shown in Figure~\ref{fig:graphs-road}. In some cases, road-level graphs are the most suitable format, e.g., when vehicles can move only through pre-defined roads.

\begin{figure*}[!htb]
\centering
\subfigure[][]{
    \includegraphics[width=0.17\textwidth]{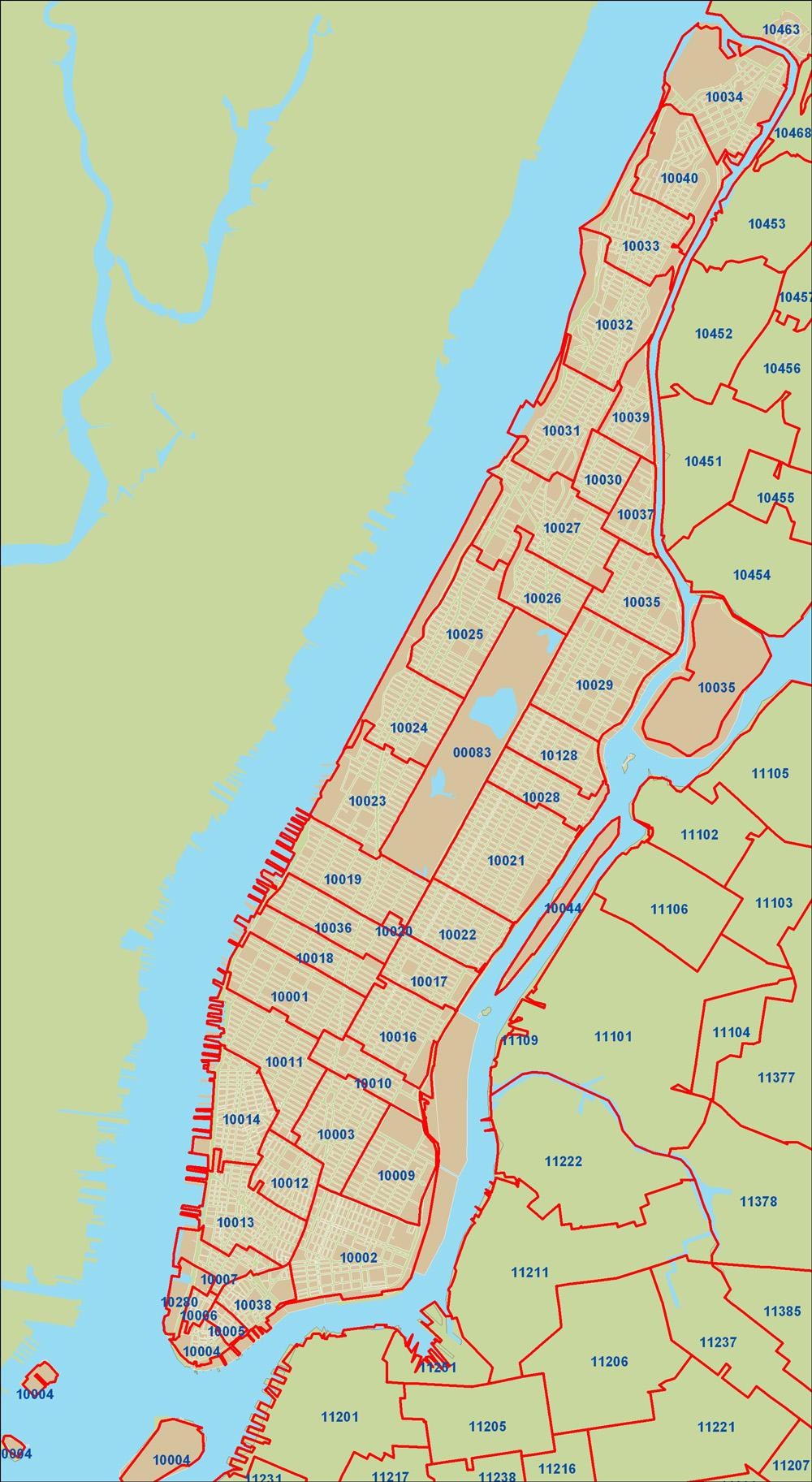}
    \label{fig:manhataan}
} 
\subfigure[][]{
    \includegraphics[width=0.75\textwidth]{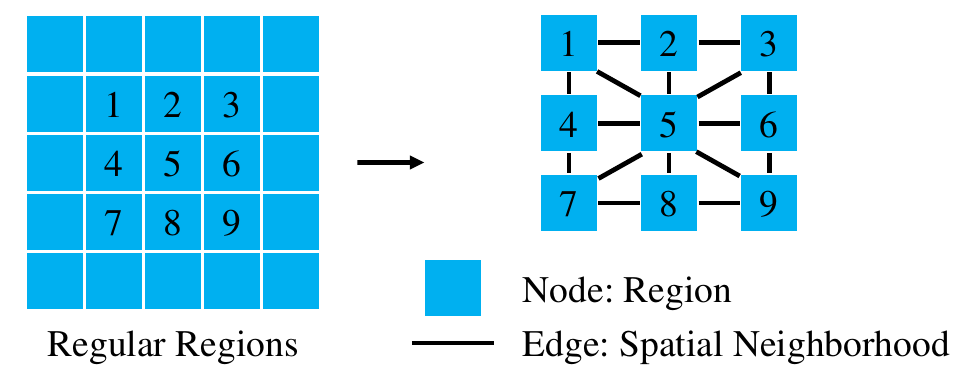}
    \label{fig:fig2b}
} 
\caption[]{The real-world case and example of region-level graphs.
    \subref{fig:manhataan} The zip codes of Manhattan where each zip code zone is a node. Source: \url{https://maps-manhattan.com/manhattan-zip-code-map}.
    \subref{fig:fig2b} The region-level graph example.}
\label{fig:graphs-region}
\end{figure*}

\textit{Region-level graphs}. These include \textit{irregular region graphs}, \textit{regular region graphs}, and \textit{OD graphs}. In both irregular and regular region graphs the nodes are regions of the city. Regular region graphs, which have grid-based partitioning, are listed separately because of their natural connection to previous widely used grid-based forecasting using CNNs, in which the grids may be seen as image pixels. Irregular region graphs include all other partitioning approaches, e.g. road based, or zip code based~\cite{ke2019predicting}. In the OD graph, the nodes are origin region - destination region pairs. In these graphs, the edges are usually defined with a spatial neighborhood or other similarities, e.g., functional similarity derived from point-of-interests (PoI) data. The real-world case and example of region-level graphs are shown in Figure~\ref{fig:graphs-region}.

\begin{figure*}[!htb]
\centering
\subfigure[][]{
    \includegraphics[width=0.5\textwidth]{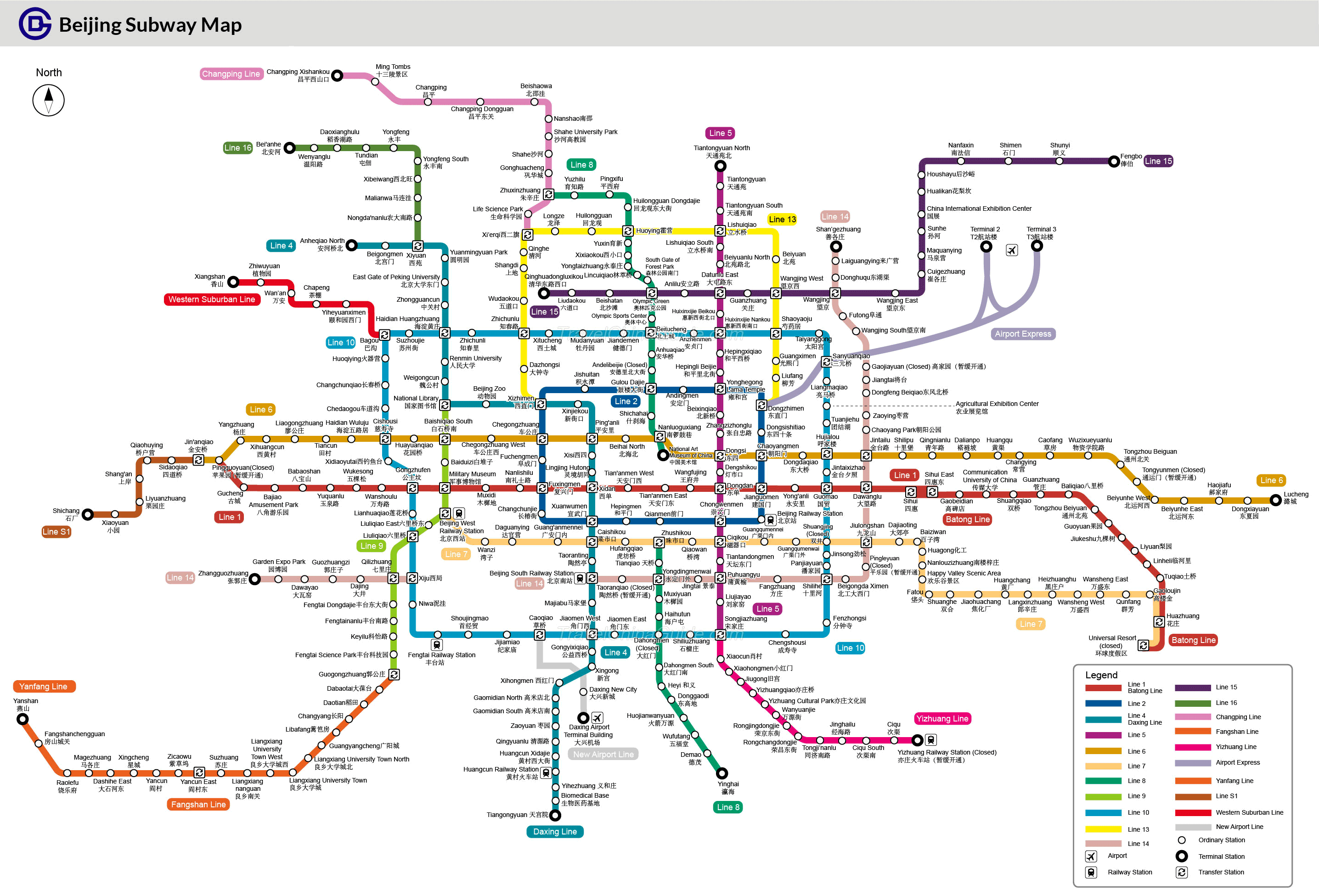}
    \label{fig:beijing}
}
\subfigure[][]{
    \includegraphics[width=0.4\textwidth]{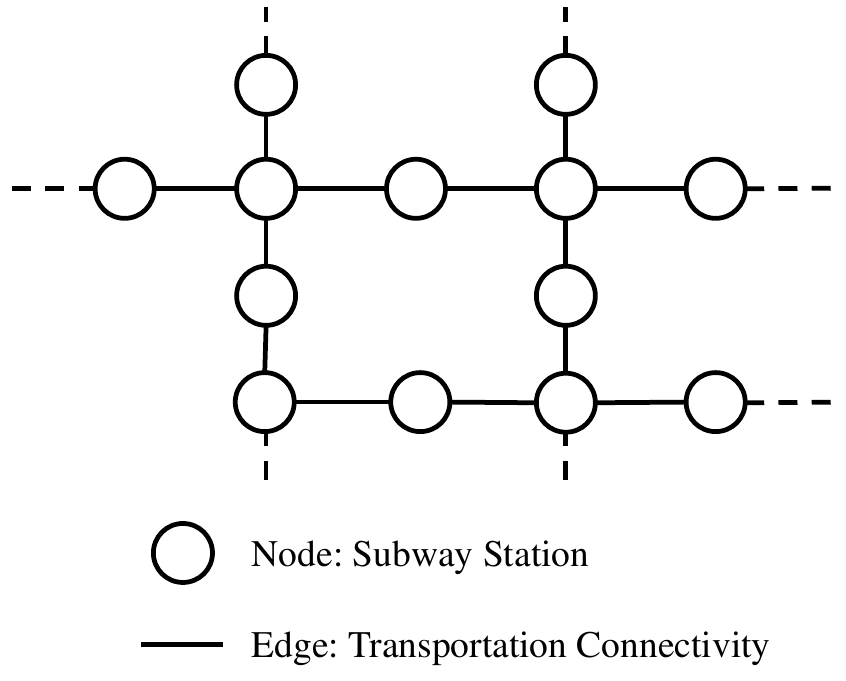}
    \label{fig:fig2c}
} 
\caption[]{The real-world case and example of station-level graphs.
    \subref{fig:beijing} The Beijing subway system where each subway station is a node. Source: \url{https://www.travelchinaguide.com/cityguides/beijing/transportation/subway.htm}.
    \subref{fig:fig2c} The station-level graph example.}
\label{fig:graphs-station}
\end{figure*}

\textit{Station-level graphs}. These include \textit{subway station graphs}, \textit{bus station graphs}, \textit{bike station graphs}, \textit{railway station graphs}, \textit{car-sharing station graphs}, \textit{parking lot graphs}, and \textit{parking block graphs}. Usually, there are natural links between stations that are used to define the edges, e.g. subway or railway lines, or the road network. The real-world case and example of station-level graphs are shown in Figure~\ref{fig:graphs-station}.

A full list of the traffic graphs used in the surveyed studies is shown in Table~\ref{tab:graphs}. Sensor graphs and road segment graphs are most frequently used because they are compatible with the available public datasets as discussed in Section~\ref{sec:resources}. It is noted that in some studies multiple graphs are used as simultaneous inputs and then fused to improve the forecasting performance~\citep{lv2020temporal, zhu2019multistep}.

\begin{landscape}
\begin{table}[!htb]
\tiny
    \centering
    \caption{Traffic graphs in the surveyed studies.}
    \label{tab:graphs}
    \begin{tabular}{|p{3cm}|p{1.5cm}|p{1.5cm}|p{12cm}|}
    \hline Graph & Node & Edge & Relevant Studies \\
    \hline
    Sensor Graph & Traffic Sensors & Road Links & \cite{li2018dcrnn_traffic, wu2019graph, xu2020ge, zheng2020gman, pan2020spatio, pan2019urban, lu2019graph, mallick2020graph, zhang2020novel, bai2020adaptive, huang2020lsgcn, zhang2020spatio, song2020spatial, xu2020spatial, wang2020traffic, chen2020gdcrn, lv2020temporal, kong2020stgat, fukuda2020short, zhang2020graphieee, boukerche2020performance, tang2020general, kang2019learning, guo2019attention, li2019hybrid, sun2020traffic, wei2020spatial, li2020multi, cao2020spectral, bing2018spatio, yu2019st, li2020traffic, yin2020multi, chen2020tssrgcn, zhang2020attention, wang2020auto, xin2020internet, xie2020sast, huang2020short, mengzhang2020spatial, tian2020st, xu2020traffic, chen2020gst, xiong2020dynamic, chen2020dynamic, tang2020dynamic, zhang2018gaan, zhang2019spatial, cirstea2019graph, ge2019temporal, ge2019traffic, shleifer2019incrementally, ge2020global, yang2020relational, zhao2019t, cui2019traffic, chen2019gated, yu20193d, wang2020forecast, cui2020graph, cui2020learning, zhou2020reinforced, cai2020traffic, zhou2020variational, wu2020connecting, chen2020multi, opolka2019spatio, mallick2020transfer, oreshkin2020fc, jia2020dynamic, sun2020constructing, guo2020short, zhang2020spatial1, feng2020dynamic, xie2020istd, park2020st, chen2020graph, lewenfus2020joint, maas2020uncertainty, li2020two, song2020graph, zhao2020attention, wang2020evaluation} \\
    \hline
    Road Segment Graph & Road Segments & Road Intersections & \cite{zhang2018kernel, guo2020optimized, pan2019urban, zhang2020novel, zhang2020urban, wang2018efficient, zhang2020spatio, lv2020temporal, zhang2019graph, zhang2020attention, qu2020modeling, guo2020dynamic, ramadan2020traffic, zhao2020spatiotemporal, zhu2020a3t, shin2020incorporating, liu2020graphsage, yu2019real, xie2019sequential, guo2019multi, diao2019dynamic, lu2019leveraging, zhang2019link, james2019online, zhang2019hybrid, lee2019ddp, yu2020forecasting, lu2020lstm, zhao2019t, cui2019traffic, zhang2019gcgan, lee2019graph, cui2020graph, cui2020learning, guo2020an, xie2020deep, zhu2020ast, zhu2020novel, fu2020bayesian, zhang2020graph, agafonov2020traffic, lu2020spatiotemporal, jepsen2019graph, jepsen2020relational, zhu2020kst, liao2018deep, guopeng2020dynamic, kim2020urban, hasanzadeh2019piecewise, fang2020constgat, dai2020hybrid, han2020congestion, hong2020heteta, chen2020multitask, yu2020deep} \\
    \hline
    Road Intersection Graph & Road Intersections & Road Segments & \cite{zhang2018kernel, wei2019dual, fang2019gstnet, zhang2020spatio, xu2019road, wu2018graph, sanchez2020gannster, james2020citywide, zhang2019multistep, lu2019leveraging, zhang2019link, bogaerts2020graph, shao2020estimation, qin2020graph} \\
    \hline
    Road Lane Graph & Road Lanes & Road Line Adjacency & \cite{wright2019neural} \\
    \hline
    Irregular Region Graph & Irregular Regions & Regional Adjacency or Virtual Edges & \cite{zhou2020exploiting, sun2020predicting, chen2020dynamic, bing2020integrating, mohanty2018graph, mohanty2020region, hu2018recurrent, li2020short, bai2019stg2seq, bai2019spatio, ke2020joint, hu2020stochastic, zheng2020spatial, davis2020grids, du2020traffic, li2020forecaster, ye2020coupled, zhang2020multi1, Liu2020stmfm} \\
    \hline
    Regular Region Graph & Regular Regions & Regional Adjacency or Virtual Edges & \cite{pan2020spatio, wang2020mtgcn, zhang2020spatial, wang2018graph, zhou2019revisiting, wang2020multi, qiu2020topological, he2020dynamic, yeghikyan2020learning, shi2020predicting, wang2019origin, shen2020ttpnet, pian2020spatial, jin2020deep, jin2020urban, geng2019multi, lee2019demand, geng2019spatiotemporal, li2020sdcn, xu2019incorporating, davis2020grids, wu2020multi, zhou2020riskoracle, zhou2020foresee, xu2020spatiotemporal} \\
    \hline
    OD Graph & OD Pair & Virtual Edges & \cite{wang2020urban, ke2019predicting} \\
    \hline
    Subway Station Graph & Subway Stations & Subway Lines & \cite{fang2019gstnet, fang2020meta, ren2019transfer, li2018graph, zhao2020spatiotemporal, han2019predicting, zhang2020deep, zhang2020multi, li2020tensor, liu2020physical, ye2020multi, ou2020stp} \\
    \hline
    Bus Station Graph & Bus Stations & Bus Lines & \cite{fang2019gstnet, fang2020meta} \\
    \hline
    Bike Station Graph & Bike Stations & Road Links & \cite{he2020towards, chai2018bike, du2020traffic, chen2020context, wang2020spatial, qin2020resgcn, xiao2020demand, yoshida2019practical, guo2019bikenet, kim2019graph, lin2018predicting} \\
    \hline
    Railway Station Graph & Railway Stations & Railway Lines & \cite{he2020gc, heglund2020railway} \\
    \hline
    Car-sharing Station Graph & Car-sharing Stations & Road Links & \cite{zhu2019multistep, luo2020d3p} \\
    \hline
    Parking Lot Graph & Parking Lots & Road Links & \cite{zhang2020semijournal, zhang2020semi} \\
    \hline
    Parking Block Graph & Parking Blocks & Road Links & \cite{yang2019deep} \\
    \hline
    \end{tabular}
\end{table}
\end{landscape}

\subsubsection{Adjacency Matrix Construction}
Adjacency matrices are seen as the key to capturing spatial dependency in traffic forecasting~\citep{ye2020build}. While nodes may be fixed by physical constraints, the user typically has control over the design of the adjacency matrix, which can even be dynamically trained from continuously evolving data. We extend the categories of adjacency matrices used in previous studies~\citep{ye2020build} and divide them into four types, namely, road-based, distance-based, similarity-based, and dynamic matrices.

\textit{Road-based Matrix}. This type of adjacency matrix relates to the road network and includes connection matrices, transportation connectivity matrices, and direction matrices. A connection matrix is a common way of representing the connectivity between nodes. It has a binary format, with an element value of 1 if connected and 0 otherwise. The transportation connectivity matrix is used where two regions are geographically distant but conveniently reachable by motorway, highway, or subway~\citep{ye2020build}. It also includes cases where the connection is measured by travel time between different nodes, e.g. if a vehicle can travel between two intersections in less than 5 minutes then there is an edge between the two intersections~\citep{wu2018graph}. The less commonly used direction matrix takes the angle between road links into consideration.

\textit{Distance-based Matrix}. This widely used matrix-type represents the spatial closeness between nodes. It contains two sub-types, namely, neighbor and distance matrices. In neighbor matrices, the element values are determined by whether the two regions share a common boundary (if connected the value is set to 1, generally, or 1/4 for grids, and 0 otherwise). In distance-based matrices, the element values are a function of geometrical distance between nodes. This distance may be calculated in various ways, e.g. the driving distance between two sensors, the shortest path length along the road~\citep{kang2019learning, lee2019ddp}, or the proximity between locations calculated by the random walk with restart (RWR) algorithm~\citep{zhang2019gcgan}. One flaw of distance-based matrices is that the fail to take into account the similarity of traffic states between long-distance nodes, and the constructed adjacency matrix is static in most cases.

\textit{Similarity-based Matrix}. This type of matrix is divided into two sub-types, namely, traffic pattern and functional similarity matrices. Traffic pattern similarity matrices represent the correlations between traffic states, e.g. similarities of flow patterns, mutual dependencies between different locations, and traffic demand correlation in different regions. Functional similarity matrices represent, for example, the distribution of different types of PoIs in different regions.

\textit{Dynamic Matrix}. This type of matrix is used when no pre-defined static matrices are used. Many studies have demonstrated the advantages of using dynamic matrices, instead of a pre-defined adjacency matrix, for various traffic forecasting problems.

A full list of the adjacency matrices applied in the surveyed studies is shown in Table~\ref{tab:matrices}. Dynamic matrices are listed at the bottom of the table, with no further subdivisions. The connection and distance matrices are the most frequently used types, because of their simple definition and representation of spatial dependency.

\begin{landscape}
\begin{table}[!htb]
\tiny
    \centering
    \caption{Adjacency matrices in the surveyed studies.}
    \label{tab:matrices}
    \begin{tabular}{|p{3cm}|p{3cm}|p{12cm}|}
    \hline Adjacency Matrix & Formula & Relevant Studies \\
    \hline
    Connection Matrix & $a_{ij} = 1$ when nodes $i$ and $j$ are connected and $a_{ij} = 0$ otherwise & \cite{zhang2018kernel, wei2019dual, xu2020ge, guo2020optimized, zhang2020urban, wang2018efficient, song2020spatial, zhang2020graphieee, xu2019road, cao2020spectral, yu2019st, chen2020tssrgcn, zhang2020attention, qu2020modeling, wang2020mtgcn, huang2020short, xiong2020dynamic, sanchez2020gannster, wang2020urban, zhang2020multi, li2020tensor, liu2020physical, ou2020stp, he2020gc, zhu2020a3t, liu2020graphsage, zhang2019multistep, yu2019real, xie2019sequential, guo2019multi, lu2019leveraging, zhang2019link, james2019online, zhang2019hybrid, zhao2019t, cui2019traffic, cui2020graph, cui2020learning, wu2020connecting, opolka2019spatio, sun2020constructing, guo2020short, xie2020deep, zhu2020ast, zhu2020novel, zhang2020graph, agafonov2020traffic, chen2020graph, lu2020spatiotemporal, bing2020integrating, zhu2020kst, fang2020constgat, shao2020estimation, shen2020ttpnet, qin2020graph, hong2020heteta, xu2019incorporating, davis2020grids, chen2020multitask, wang2020spatial, zhou2020riskoracle, yu2020deep, Liu2020stmfm, zhang2020semijournal, zhang2020semi, heglund2020railway, yin2020multi, zhang2020deep} \\
    \hline
    Transportation Connectivity Matrix & $a_{ij} = 1$ when one can travel from node $i$ to node $j$ and $a_{ij} = 0$ otherwise & \cite{pan2020spatio, pan2019urban, lv2020temporal, wu2018graph, ye2020multi, geng2019multi, geng2019spatiotemporal, luo2020d3p, wright2019neural} \\
    \hline
    Direction Matrix & $a_{ij} = $ the angle between two road segments & \cite{shin2020incorporating, lee2019ddp, lee2019graph} \\
    \hline
    Neighbor Matrix & $a_{ij} = 1$ when nodes $i$ and $j$ are neighbors and $a_{ij} = 0$ otherwise & \cite{wang2018graph, yeghikyan2020learning, shi2020predicting, wang2019origin, hu2018recurrent, geng2019multi, lee2019demand, ke2020joint, ke2019predicting, hu2020stochastic, zheng2020spatial, yoshida2019practical} \\
    \hline
    Distance Matrix & $a_{ij} = d_{ij}$ and $d_{ij}$ is some distance between nodes $i$ and $j$ & \cite{li2018dcrnn_traffic, zheng2020gman, pan2020spatio, pan2019urban, lu2019graph, mallick2020graph, huang2020lsgcn, xu2020spatial, wang2020traffic, boukerche2020performance, kang2019learning, sun2020traffic, wei2020spatial, bing2018spatio, li2020traffic, chen2020tssrgcn, wang2020auto, xin2020internet, xie2020sast, mengzhang2020spatial, tian2020st, xu2020traffic, chen2020gst, zhou2020exploiting, chen2020dynamic, he2020dynamic, ren2019transfer, zhu2019multistep, he2020towards, chai2018bike, shin2020incorporating, zhang2018gaan, ge2019temporal, ge2019traffic, lee2019ddp, shleifer2019incrementally, ge2020global, yang2020relational, chen2019gated, zhang2019gcgan, lee2019graph, bogaerts2020graph, wang2020forecast, guo2020an, zhou2020reinforced, cai2020traffic, zhou2020variational, chen2020multi, mallick2020transfer, jia2020dynamic, zhang2020spatial1, feng2020dynamic, xie2020istd, li2020two, song2020graph, zhao2020attention, kim2020urban, mohanty2018graph, mohanty2020region, jin2020deep, li2020short, jin2020urban, geng2019spatiotemporal, ke2020joint, li2020sdcn, ke2019predicting, luo2020d3p, chen2020context, xiao2020demand, guo2019bikenet, kim2019graph, lin2018predicting, yang2019deep, wang2020evaluation, xu2020spatiotemporal} \\
    \hline
    Traffic Pattern Similarity Matrix & $a_{ij} = $ the correlation coefficient of historical traffic states of nodes $i$ and $j$ & \cite{lv2020temporal, mengzhang2020spatial, xu2020traffic, zhou2020exploiting, sun2020predicting, wang2020multi, he2020dynamic, ren2019transfer, han2019predicting, liu2020physical, he2020towards, chai2018bike, lu2020spatiotemporal, lewenfus2020joint, dai2020hybrid, han2020congestion, jin2020deep, li2020short, jin2020urban, bai2019stg2seq, bai2019spatio, li2020sdcn, ke2019predicting, chen2020context, wang2020spatial, yoshida2019practical, kim2019graph, lin2018predicting, zhou2020foresee} \\
    \hline
    Functional Similarity Matrix & $a_{ij} = $ the correlation coefficient of POI distributions in regions $i$ and $j$ & \cite{lv2020temporal, he2020dynamic, shi2020predicting, zhu2019multistep, ge2019temporal, ge2019traffic, ge2020global, jin2020deep, geng2019multi, geng2019spatiotemporal, ke2019predicting, luo2020d3p, zhang2020multi1} \\
    \hline
    Dynamic Matrix & N/A & \cite{wu2019graph, bai2020adaptive, fang2019gstnet, zhang2020spatio, chen2020gdcrn, kong2020stgat, tang2020general, guo2019attention, li2019hybrid, zhang2019graph, li2020multi, guo2020dynamic, zhang2020spatial, peng2020spatial, zhou2019revisiting, shi2020predicting, li2018graph, tang2020dynamic, zhang2019spatial, diao2019dynamic, yu2020forecasting, fu2020bayesian, maas2020uncertainty, li2020short, du2020traffic, li2020forecaster, wu2020multi, ye2020coupled} \\
    \hline
    \end{tabular}
\end{table}
\end{landscape}

\subsection{Graph Neural Networks}
Previous neural networks, e.g. fully-connected neural networks (FNNs), CNNs, and RNNs, could only be applied to Euclidean data (i.e. images, text, and videos). As a type of neural network which directly operates on a graph structure, GNNs have the ability to capture complex relationships between objects and make inferences based on data described by graphs. GNNs have been proven effective in various node-level, edge-level, and graph-level prediction tasks~\citep{jiang2022graph}. As mentioned in Section~\ref{sec:related}, GNNs are currently considered the state-of-the-art techniques for traffic forecasting problems. GNNs can be roughly divided into four types, namely, recurrent GNNs, convolutional GNNs, graph autoencoders, and spatiotemporal GNNs~\citep{wu2020comprehensive}. Because traffic forecasting is a spatiotemporal problem, the GNNs used in this field can all be categorized as the spatiotemporal GNNs. However, certain components of the other types of GNNs have also been applied in the surveyed traffic forecasting studies.

To give the mathematical formulation of GCN, we further introduce some notations. Give a graph $G=(V, E, A)$, $\mathcal{N}(v_i)$ is defined as the neighbor node set of a single node $v_i$. $\mathbf{D}$ is defined as the degree matrix, of which each element is $\mathbf{D}_{ii}=\|\mathcal{N}(v_i)\|$. $\mathbf{L} = \mathbf{D} - \mathbf{A}$ is defined as the Laplacian matrix of an undirected graph and $\tilde{\mathbf{L}} = \mathbf{I}_N - \mathbf{D}^{-\frac{1}{2}} \mathbf{A} \mathbf{D}^{-\frac{1}{2}}$ is defined as the normalized Laplacian matrix, where $\mathbf{I}_N$ is the identity matrix with size $N$. Without considering the time step index, the node feature matrix of a graph is simplified as $\mathbf{X} \in {R}^{N \times d}$, where $N$ is the node number and $d$ is the dimension of the node feature vector as before. The basic notations used in this survey is summarized in Table~\ref{tab:notations}.

\begin{table}[!htb]
\tiny
    \centering
    \caption{Basic notations used in this study.}
    \label{tab:notations}
    \begin{tabular}{|l|l|}
    \hline
    Symbol & Description \\
    \hline
    $G$ & Graph \\
    $V$ & Node set \\
    $E$ & Edge set \\
    $A$ & Adjacency matrix \\
    $\chi_t$ or $\mathbf{X}$ & Node feature matrix w/o time step index $t$ \\
    $N$ & Node number \\
    $d$ & Node feature dimension \\
    $\mathcal{N}(v_i)$ & Neighbor node set of a single node $v_i$ \\
    $\mathbf{D}$ & Degree matrix \\
    $\mathbf{L}$ & Laplacian matrix \\
    $\tilde{\mathbf{L}}$ & Normalized Laplacian matrix \\
    $\mathbf{I}_N$ & Identity matrix with size $N$ \\
    \hline
    \end{tabular}
\end{table}

When extending the convolution operation from Euclidean data to non-Euclidean data, the basic idea of GNNs is to learn a function mapping for a node to aggregate its own features and the features of its neighbors to generate a new representation. GCNs are spectral-based convolutional GNNs, in which the graph convolutions are defined by introducing filters from graph signal processing in the spectral domain, e.g., the Fourier domain. The graph Fourier transform is firstly used to transform the graph signal to the spectral domain and the inverse graph Fourier transform is further used to transform the result after the convolution operation back. Several spectral-based GCNs are introduced in the literature. Spectral convoluted neural networking~\citep{bruna2014spectral} assumes that the filter is a set of learnable parameters and considers graph signals with multiple channels. GNN~\citep{henaff2015deep} introduces a parameterization with smooth coefficients and makes the spectral filters spatially localized. Chebyshev’s spectral CNN (ChebNet)~\citep{defferrard2016convolutional} leverages a truncated expansion in terms of Chebyshev polynomials up to $K$th order to approximate the diagonal matrix.

GCN~\citep{kipf2017semi} is a first-order approximation of ChebNet, which approximates the filter using the Chebyshev polynomials of the diagonal matrix of eigenvalues. To avoid overfitting, $K=1$ is used in GCN. Formally, the graph convolution operation $*G$ in GCN is defined as follows:
\begin{equation}
\mathbf{X}_{*G} = \mathbf{W} (\mathbf{I}_N + \mathbf{D}^{-\frac{1}{2}} \mathbf{A} \mathbf{D}^{-\frac{1}{2}}) \mathbf{X}
\end{equation}
\noindent where $\mathbf{W}$ is a learnable weight matrix, i.e., the model parameters. While in practice, the graph convolution operation is further developed in order to alleviate the potential gradient explosion problem as follows:
\begin{equation}
\mathbf{X}_{*G} = \mathbf{W} (\tilde{\mathbf{D}}^{-\frac{1}{2}} \tilde{\mathbf{A}} \tilde{\mathbf{D}}^{-\frac{1}{2}}) \mathbf{X}
\end{equation}
\noindent where $\tilde{\mathbf{A}} = \mathbf{A} + \mathbf{I}_N$ and $\tilde{\mathbf{D}}_{ii} = \sum_{j}{\tilde{\mathbf{A}}_{ij}}$.

The alternative approach is spatial-based convolutional GNNs, in which the graph convolutions are defined by information propagation. Diffusion graph convolution (DGC)~\citep{atwood2016diffusion}, message passing neural network (MPNN)~\citep{gilmer2017neural}, GraphSAGE~\citep{hamilton2017inductive}, and graph attention network (GAT)~\citep{velivckovic2018graph} all follow this approach. The graph convolution is modeled as a diffusion process with a transition probability from one node to a neighboring node in DGC. An equilibrium is expected to be obtained after several rounds of information transition. The general framework followed is a message passing network, which models the graph convolutions as an information-passing process from one node to another connected node directly. To alleviate the computation problems caused by a large number of neighbors, sampling is used to obtain a fixed number of neighbors in GraphSAGE. Lastly, without using a predetermined adjacency matrix, the attention mechanism is used to learn the relative weights between two connected nodes in GAT.

MPNN uses message passing functions to unify different spatial-based variants. MPNN operates in two stages, namely, a message passing phase and a readout phase. The message passing phase is defined as follows:
\begin{equation}
\mathbf{m}_{v_i}^{(t)} = \sum_{v_j \in \mathcal{N}{(v_i)}} \mathcal{M}^{(t)} (\mathbf{X}_i^{(t-1)}, \mathbf{X}_j^{(t-1)}, \mathbf{e}_{ij})
\end{equation}
\noindent where $\mathbf{m}_{v_i}^{(t)}$ is the message aggregated from the neighbors of node $v_i$, $\mathcal{M}^{(t)}(\cdot)$ is the aggregation function in the $t$-th iteration, $\mathbf{X}_i^{(t)}$ is the hidden state of node $v_i$ in the $t$-th iteration, and $\mathbf{e}_{ij}$ is the edge feature vector between node $v_i$ and node $v_j$. 

The readout phase is defined as follows:
\begin{equation}
\mathbf{X}_i^{(t)} = \mathcal{U}^{(t)} (\mathbf{X}_i^{(t-1)},\mathbf{m}_{v_i}^{(t)})
\end{equation}
\noindent where $\mathcal{U}^{(t)}(\cdot)$ is the readout function in the $t$-th iteration.

In GAT~\citep{velivckovic2018graph}, the attention mechanism~\citep{vaswani2017attention} is incorporated into the propagation step and the multi-head attention mechanism is further utilized with the aim of stabilizing the learning process. The specific operation is defined as follows:
\begin{equation}
\mathbf{X}_i^{(t)} = \|_k \sigma (\sum_{j \in \mathcal{N}{(v_i)}} \alpha^k (\mathbf{X}_i^{(t-1)}, \mathbf{X}_j^{(t-1)}) \mathbf{W}^{(t-1)} \mathbf{X}_j^{(t-1)})
\end{equation}
\noindent where $\|$ is the concatenation operation, $\sigma$ is the activation method, $\alpha^{k}(\cdot)$ is the $k$-th attention mechanism.

\begin{figure*}[!htb]
\centering
\subfigure[][]{
    \includegraphics[width=\textwidth]{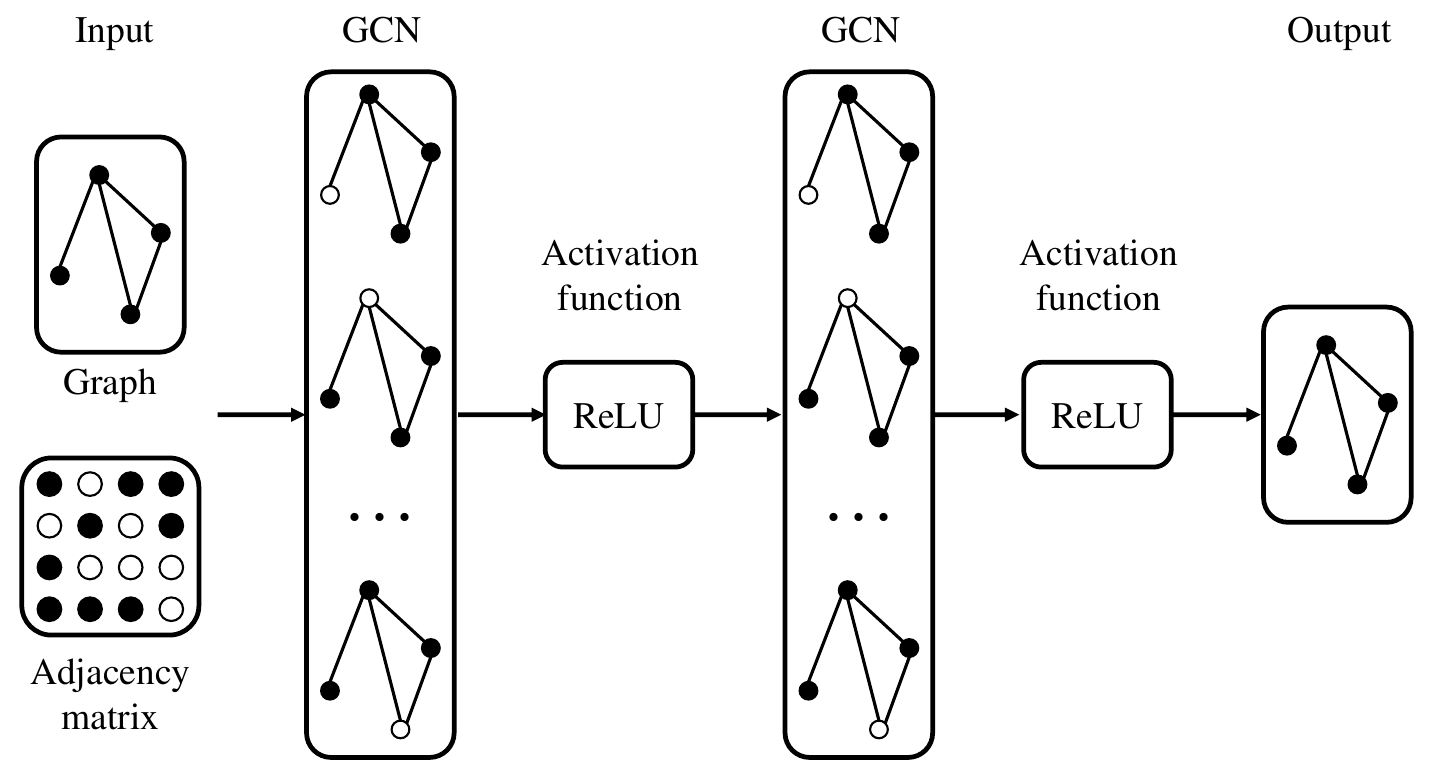}
    \label{fig:fig3a}
} \\
\subfigure[][]{
    \includegraphics[width=\textwidth]{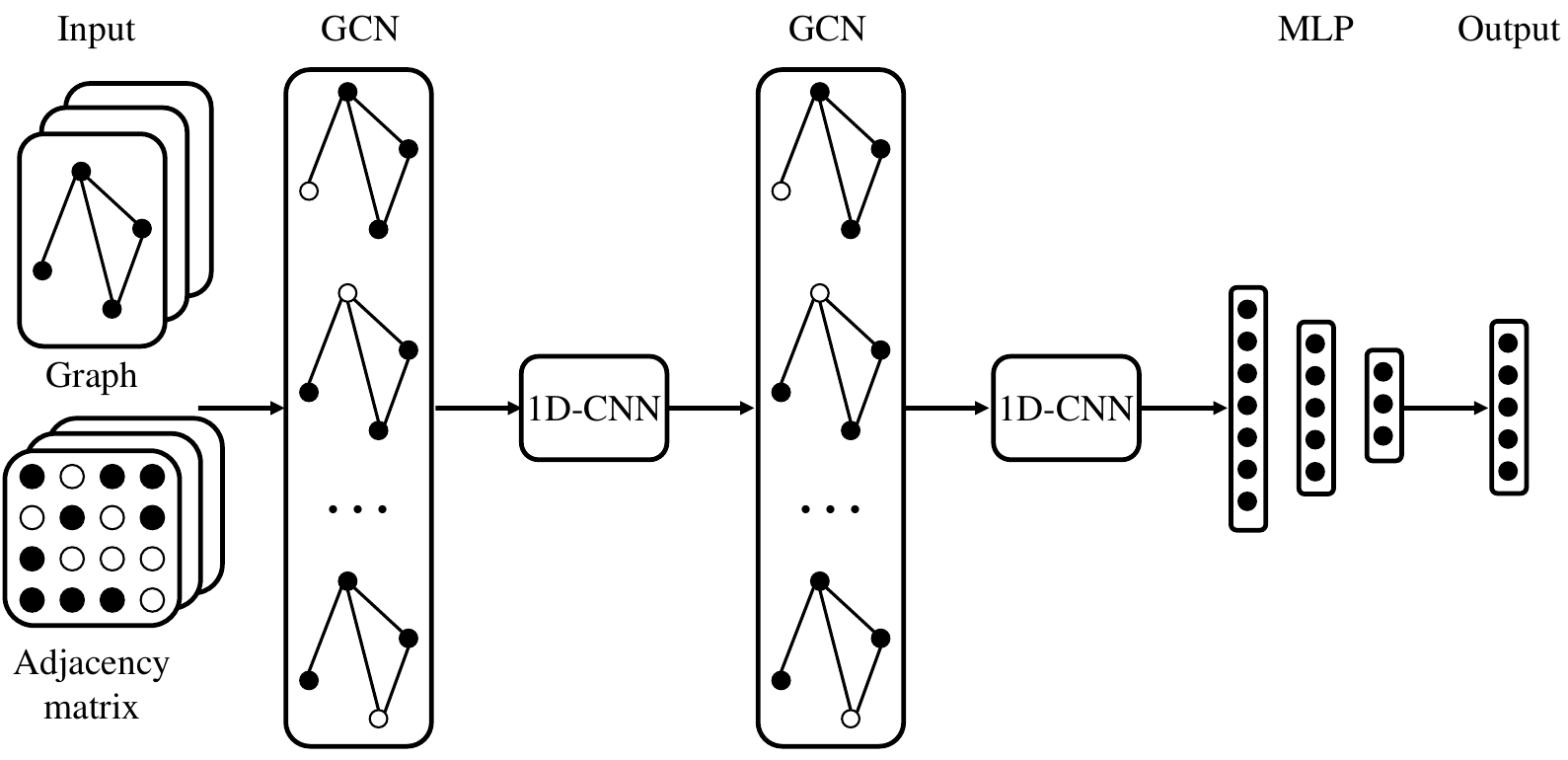}
    \label{fig:fig3b}
}
\caption[]{A comparison between a two-layer GCN model and a typical spatiotemporal GNN structure (1D-CNN+GCN as an example). Adapted from~\cite{survey2021wang}.
    \subref{fig:fig3a} a two-layer GCN model;
    \subref{fig:fig3b} a typical spatiotemporal GNN structure.}
\label{fig:fig3}
\end{figure*}

A general spatiotemporal GNN structure is shown in Figure~\ref{fig:fig3b}, in which GCN is used to capture the spatial dependency and 1D-CNN is used to capture the temporal dependency. Both GCN and 1D-CNN components can be replaced with other structures for other spatiotemporal GNNs. A multilayer perceptron (MLP) component is used to generate the desired output. As for comparison, a two-layer GCN is also shown in Figure~\ref{fig:fig3a}, in which only the spatial dependency is concerned.

Spatiotemporal GNNs can be further categorized based on the approach used to capture the temporal dependency in particular. Most of the relevant studies in the literature can be split into two types, namely, \textit{RNN-based} and \textit{CNN-based} spatiotemporal GNNs~\citep{wu2020comprehensive}. The RNN-based approach is used in~\cite{li2018dcrnn_traffic, guo2020optimized, pan2020spatio, pan2019urban, lu2019graph, mallick2020graph, zhang2020novel, zhang2020urban, bai2020adaptive, huang2020lsgcn, wang2018efficient, wang2020traffic, lv2020temporal, fukuda2020short, zhang2020graphieee, boukerche2020performance, kang2019learning, li2019hybrid, xu2019road, wu2018graph, wei2020spatial, li2020multi, yu2019st, yin2020multi, xin2020internet, qu2020modeling, huang2020short, guo2020dynamic, fang2020meta, mengzhang2020spatial, chen2020gst, ramadan2020traffic, zhou2020exploiting, wang2018graph, peng2020spatial, zhou2019revisiting, wang2020multi, qiu2020topological, shi2020predicting, wang2020urban, wang2019origin, zhang2020deep, liu2020physical, ye2020multi, zhu2019multistep, chai2018bike, he2020gc, zhu2020a3t, zhang2018gaan, zhang2019multistep, xie2019sequential, zhang2019spatial, guo2019multi, cirstea2019graph, lu2019leveraging, zhang2019hybrid, lu2020lstm, zhao2019t, cui2019traffic, chen2019gated, zhang2019gcgan, bogaerts2020graph, cui2020learning, zhou2020reinforced, mallick2020transfer, sun2020constructing, xie2020deep, zhu2020ast, zhu2020novel, fu2020bayesian, chen2020graph, lewenfus2020joint, zhu2020kst, liao2018deep, zhao2020attention, guopeng2020dynamic, shao2020estimation, shen2020ttpnet, mohanty2018graph, mohanty2020region, hu2018recurrent, pian2020spatial, jin2020urban, geng2019spatiotemporal, bai2019spatio, li2020sdcn, ke2019predicting, hu2020stochastic, xu2019incorporating, davis2020grids, chen2020multitask, du2020traffic, wu2020multi, ye2020coupled, luo2020d3p, chen2020context, wang2020spatial, xiao2020demand, guo2019bikenet, lin2018predicting, zhou2020foresee, Liu2020stmfm, zhang2020semijournal, yang2019deep, zhang2020semi, wang2020evaluation, wright2019neural}; while the CNN-based approach is used in~\cite{wu2019graph, fang2019gstnet, zhang2020spatio, xu2020spatial, chen2020gdcrn, kong2020stgat, tang2020general, guo2019attention, sun2020traffic, bing2018spatio, li2020traffic, wang2020auto, tian2020st, chen2020dynamic, zhao2020spatiotemporal, zhang2020multi, ou2020stp, tang2020dynamic, diao2019dynamic, lee2019ddp, lee2019graph, wang2020forecast, wu2020connecting, guo2020short, zhang2020spatial1, feng2020dynamic, zhang2020graph, xie2020istd, lu2020spatiotemporal, maas2020uncertainty, li2020two, song2020graph, dai2020hybrid, hong2020heteta, zheng2020spatial, zhou2020riskoracle, yu2020deep, xu2020spatiotemporal, heglund2020railway}. 

With the recent expansion of relevant studies, we add two sub-types of spatiotemporal GNNs in this survey, namely, \textit{attention-based} and \textit{FNN-based}. Attention mechanism is firstly proposed to memorize long source sentences in neural machine translation~\citep{vaswani2017attention}. Then it is used for temporal forecasting problems. As a special case, Transformer is built entirely upon attention mechanisms, which makes it possible to access any part of a sequence regardless of its distance to the target~\citep{xie2020sast, cai2020traffic, jin2020deep, li2020forecaster}. The attention-based approaches are used in~\cite{zheng2020gman, zhang2020attention, wang2020mtgcn, xie2020sast, cai2020traffic, zhou2020variational, chen2020multi, park2020st, fang2020constgat, jin2020deep, bai2019stg2seq, li2020forecaster, zhang2020multi1}, while the simpler FNN-based approach is used in~\cite{zhang2018kernel, wei2019dual, song2020spatial, cao2020spectral, chen2020tssrgcn, zhang2020spatial, sun2020predicting, he2020dynamic, yeghikyan2020learning, ren2019transfer, li2018graph, han2019predicting, he2020towards, zhang2019link, ge2019temporal, ge2019traffic, yu2020forecasting, ge2020global, yu20193d, guo2020an, agafonov2020traffic, geng2019multi, qin2020resgcn, kim2019graph}. Apart from using neural networks to capture temporal dependency, other techniques that have also been combined with GNNs include autoregression~\citep{lee2019demand}, Markov processes~\citep{cui2020graph}, and Kalman filters~\citep{xiong2020dynamic}.

Among different approaches for temporal modeling, RNNs suffer from time-consuming iterations and gradient vanishing or explosion problem with long sequences. CNNs demonstrate their superiority in terms of simple structure, parallel computing and stable gradients. As for the traffic problems, the spatial and temporal dependencies are closely intertwined in reality. For example, it is argued that the historical observations in different locations at different times have varying impacts on central region in the future~\citep{guo2019attention}. Some efforts are put to jointly modeling the potential interaction between spatial and temporal features and one promising direction is the incorporate of the graph convolution operations into RNNs to capture spatial-temporal correlations~\citep{yu2019st, zhou2019revisiting, chen2019gated, liu2020physical, chen2020multi, guo2020optimized}. For example, the localized spatio-temporal correlation information is extracted simultaneously with the adjacency matrix of localized spatio-temporal graph in~\cite{song2020spatial}, in which a localized spatio-temporal graph that includes both temporal and spatial attributes is constructed first and a spatial-based GCN method is applied then.

\begin{figure*}[!htb]
    \centering
    \includegraphics[width=\textwidth]{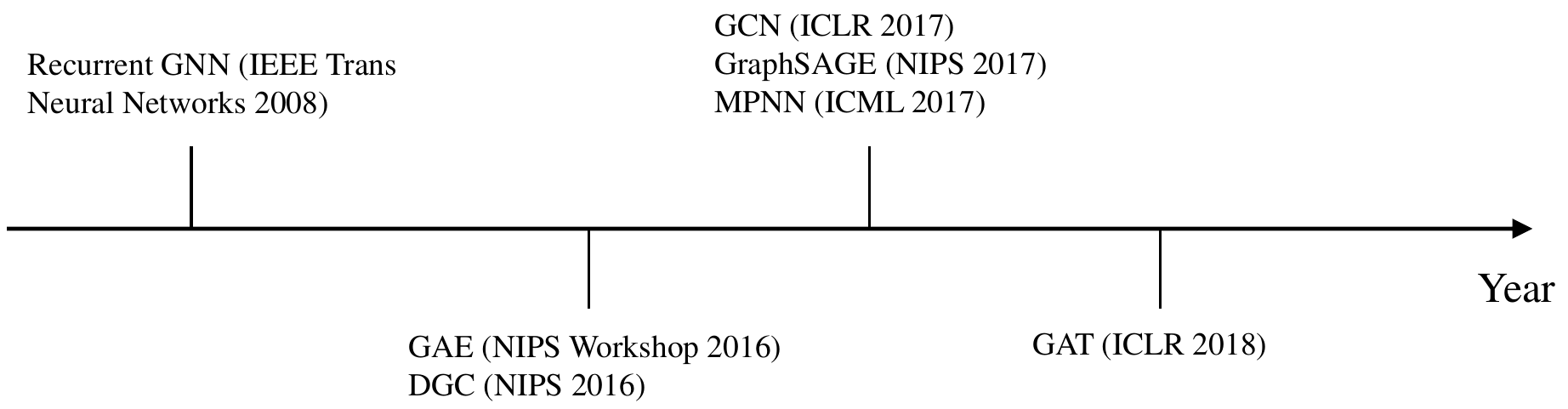}
    \caption{The relevant graph neural networks in this survey.}
    \label{fig:gnns}
\end{figure*}

Of the additional GNN components adopted in the surveyed studies, convolutional GNNs are the most popular, while recurrent GNN~\citep{scarselli2008graph} and Graph Auto-Encoder (GAE)~\citep{kipf2016variational} are used less frequently. We further categorize convolutional GNNs into the following five types: (1) GCN~\citep{kipf2017semi}, (2) DGC~\citep{atwood2016diffusion}, (3) MPNN~\citep{gilmer2017neural}, (4) GraphSAGE~\citep{hamilton2017inductive}, and (5) GAT~\citep{velivckovic2018graph}. These relevant graph neural networks are listed chronologically in Figure~\ref{fig:gnns}. While different GNNs can be used for traffic forecasting, a general design pipeline is proposed in~\citep{zhou2018graph} and suggested for future studies as follows:
\begin{enumerate}
    \item Find graph structure. As discussed in Section IV, different traffic graphs are available.
    \item Specify graph type and scale. The graphs can be further classified into different types if needed, e.g., directed/undirected graphs, homogeneous/heterogeneous graphs, static/dynamic graphs. For most cases in traffic forecasting, the graphs of the same type are used in a single study. As for the graph scale, the graphs in the traffic domain are not as large as those for the social networks or academic networks with millions of nodes and edges.
    \item Design loss function. The training setting usually follows the supervised approach, which means the GNN-based models are firstly trained on a training set with labels and then evaluated on a test set. The forecasting task is usually designed as the node-level regression problem. Based on these considerations, the proper loss function and evaluation metrics can be chosen, e.g., root mean square error (RMSE), mean absolute error (MAE) and mean absolute percentage error (MAPE).
    \item Build model using computational modules. The GNNs discussed in this section are exactly those which have already been used as computational modules to build forecasting models in the surveyed studies.
\end{enumerate}

A full list of the GNN components used in the surveyed studies is shown in Table~\ref{tab:gnns}. Currently, the most widely used GNN is the GCN. However, we also notice a growing trend in the use of GAT in traffic forecasting.

\begin{table}[!htb]
\tiny
    \centering
    \caption{GNNs in the surveyed studies.}
    \label{tab:gnns}
    \begin{tabular}{|l|p{9cm}|}
    \hline GNN & Relevant Studies \\
    \hline
    Recurrent GNN & \cite{wang2018efficient, wang2018graph, lu2019leveraging, lu2020lstm} \\
    \hline
    GAE & \cite{xu2020ge, xu2019road, opolka2019spatio, shen2020ttpnet} \\
    \hline
    GCN & \cite{wu2019graph, zhang2018kernel, guo2020optimized, lu2019graph, zhang2020novel, zhang2020urban, bai2020adaptive, fang2019gstnet, zhang2020spatio, song2020spatial, xu2020spatial, wang2020traffic, lv2020temporal, boukerche2020performance, tang2020general, guo2019attention, li2019hybrid, zhang2019graph, sun2020traffic, li2020multi, cao2020spectral, bing2018spatio, yu2019st, li2020traffic, chen2020tssrgcn, zhang2020attention, wang2020auto, xin2020internet, qu2020modeling, wang2020mtgcn, huang2020short, guo2020dynamic, fang2020meta, mengzhang2020spatial, xu2020traffic, chen2020gst, xiong2020dynamic, ramadan2020traffic, zhou2020exploiting, sun2020predicting, peng2020spatial, zhou2019revisiting, wang2020multi, qiu2020topological, he2020dynamic, yeghikyan2020learning, shi2020predicting, wang2020urban, ren2019transfer, li2018graph, zhao2020spatiotemporal, han2019predicting, zhang2020deep, zhang2020multi, liu2020physical, ye2020multi, zhu2019multistep, chai2018bike, he2020gc, zhu2020a3t, tang2020dynamic, james2020citywide, zhang2018gaan, zhang2019multistep, yu2019real, guo2019multi, diao2019dynamic, zhang2019link, james2019online, ge2019temporal, ge2019traffic, zhang2019hybrid, lee2019ddp, yu2020forecasting, ge2020global, zhao2019t, cui2019traffic, zhang2019gcgan, yu20193d, lee2019graph, bogaerts2020graph, cui2020graph, cui2020learning, guo2020an, cai2020traffic, wu2020connecting, chen2020multi, jia2020dynamic, sun2020constructing, xie2020deep, zhu2020ast, feng2020dynamic, zhu2020novel, fu2020bayesian, agafonov2020traffic, chen2020graph, lu2020spatiotemporal, jepsen2019graph, jepsen2020relational, bing2020integrating, lewenfus2020joint, zhu2020kst, liao2018deep, maas2020uncertainty, li2020two, song2020graph, zhao2020attention, guopeng2020dynamic, shao2020estimation, dai2020hybrid, mohanty2018graph, mohanty2020region, qin2020graph, han2020congestion, hong2020heteta, hu2018recurrent, li2020short, jin2020urban, geng2019multi, bai2019stg2seq, geng2019spatiotemporal, bai2019spatio, ke2020joint, li2020sdcn, ke2019predicting, hu2020stochastic, zheng2020spatial, davis2020grids, chen2020multitask, du2020traffic, li2020forecaster, ye2020coupled, luo2020d3p, chen2020context, wang2020spatial, qin2020resgcn, xiao2020demand, yoshida2019practical, guo2019bikenet, kim2019graph, lin2018predicting, zhou2020riskoracle, yu2020deep, zhang2020multi1, zhou2020foresee, Liu2020stmfm, zhang2020semijournal, yang2019deep, zhang2020semi, xu2020spatiotemporal, heglund2020railway} \\
    \hline
    DGC & \cite{li2018dcrnn_traffic, mallick2020graph, chen2020gdcrn, fukuda2020short, ou2020stp, chen2019gated, wang2020forecast, zhou2020reinforced, zhou2020variational, mallick2020transfer, xie2020istd, kim2020urban, wang2020evaluation} \\
    \hline
    MPNN & \cite{wei2019dual, xu2020spatial, wang2019origin} \\
    \hline
    GraphSAGE & \cite{liu2020graphsage} \\
    \hline
    GAT & \cite{zheng2020gman, pan2020spatio, pan2019urban, huang2020lsgcn, kong2020stgat, zhang2020graphieee, tang2020general, kang2019learning, wu2018graph, wei2020spatial, yin2020multi, xie2020sast, zhang2020spatial, tian2020st, he2020towards, tang2020dynamic, zhang2019spatial, cirstea2019graph, yang2020relational, guo2020short, zhang2020spatial1, zhang2020graph, park2020st, song2020graph, fang2020constgat, pian2020spatial, jin2020deep, xu2019incorporating, wu2020multi, wright2019neural} \\
    \hline
    \end{tabular}
\end{table}

During the process of customizing GNNs for traffic forecasting, some classical models stand out in the literature. The most famous one is diffusion convolutional recurrent neural network (DCRNN)~\citep{li2018dcrnn_traffic}, which uses diffusion graph convolutional networks and RNN to learn the representations of spatial dependencies and temporal relations. DCRNN was originally proposed for traffic speed forecasting and is now widely used as a baseline. To create the traffic graph, the adjacency matrix is defined as the thresholded pairwise road network distances. Compared with other graph convolutional models that can only operate on undirected graphs, e.g., ChebNet, DCRNN introduces the diffusion convolution (DC) operation for directed graph and is more suitable for transportation scenarios, which is defined as follows:
\begin{equation}
\mathbf{X}_{*DC} = \sum_{k=0}^{K-1} (\theta_{k, 1}(D_O^{-1} A)^k) + \theta_{k, 2}(D_I^{-1} A^T)^k) \mathbf{X}
\end{equation}
\noindent where $\mathbf{X} \in {R}^{N \times d}$ is the node feature matrix, $A$ is the adjacency matrix, $D_O$ and $D_I$ are diagonal out-degree and in-degree matrices, $\theta_{k, 1}$ and $\theta_{k, 2}$ are model parameters, $K$ is the number of diffusion steps. By defining and using out-degree and in-degree matrices, DCRNN models the bidirectional diffusion process to capture the influence of both upstream and downstream traffic. While DCRNN is a strong baseline, it is not suitable or desirable for the undirected graph cases. Then DCRNN is extended with a stronger learning ability in graph GRU in~\cite{zhang2018gaan}, in which a unified method for constructing an RNN based on an arbitrary graph convolution operator is proposed, instead of the single RNN model used in DCRNN.

Spatio-temporal graph convolutional network (STGCN)~\citep{bing2018spatio} stacks multiple spatio-temporal convolution blocks and each block concatenate two temporal convolution and one graph convolution layer. ChebNet is chosen as the graph convolution operator in STGCN, after a comparison with its first-order approximation. The usage of temporal convolution layers instead of RNNs for temporal modeling accelerates the training phase of STGCN. Attention based Spatio-temporal graph convolutional network (ASTGCN)~\citep{guo2019attention} further introduces two attention layers in STGCN to capture the dynamic correlations in spatial dimension and temporal dimension, respectively.

Graph WaveNet~\citep{wu2019graph} constructs a self-adaptive matrix to uncover unseen graph structures automatically from the data and WaveNet, which is based on causal convolutions, is used to learn temporal relations. However, the self-adaptive matrix in Graph WaveNet is fixed after training, which is unable to be adjusted dynamically with the data characteristics.

\section{Open Data and Source Codes}
\label{sec:resources}
In this section, we summarize the open data and source code used in the surveyed papers. These open data are suitable for GNN-related studies with graph structures discussed in Section IV, which can be used to formulate different forecasting problems in Section III. We also list the GNN-related code resources for those who want to replicate the previous GNN-based solutions as baselines in the follow-up studies.

\subsection{Open Data}
\label{sec:data}
We categorize the data used in the surveyed studies into three major types, namely, graph-related data, historical traffic data, and external data. Graph-related data refer to those data which exhibit a graph structure in the traffic domain, i.e., transportation network data. Historical traffic data refer to those data which record the historical traffic states, usually in different locations and time points. We further categorize the historical traffic data into sub-types as follows. External data refer to the factors that would affect the traffic states, i.e., weather data and calendar data. Some of these data can be used in the graph-based modeling directly, while the others may require some pre-processing steps before being Incorporated into GNN-based models.

\textit{Transportation Network Data}. These data represent the underlying transportation infrastructure, e.g., road, subway, and bus networks. They can be obtained from government transportation departments or extracted from online map services, e.g., OpenStreetMap. Based on their topology structure, these data can be used to build the graphs directly, e.g., the road segments or the stations are nodes and the road intersections or subway links are the edges. While this modeling approach is straightforward, the disadvantage is that only static graphs can be built from transportation network data.

\textit{Traffic Sensor Data}. Traffic sensors, e.g. loop detectors, are installed on roads to collect traffic information, e.g., traffic volume or speed. This type of data is widely used for traffic prediction, especially road traffic flow and speed prediction problems. For graph-based modeling, each sensor can be used as a node, with road connections as the edges. One advantage of using traffic sensor data for graph-based modeling is that the captured traffic information can be used directly as the node attributes, with little pre-processing overhead. One exception is that the sensors are prone to hardware faults, which causes the missing data or data noise problems and requires corresponding pre-processing techniques, e.g., data imputation and denoising methods. Another disadvantage of using traffic sensor data for graph-based modeling is that the traffic sensors can only be installed in a limited number of locations for a series of reasons, e.g., installation cost. With this constraint, only the part of the road networks with traffic sensors can be incorporated into a graph, while the uncovered areas are neglected.

\textit{GPS Trajectory Data}. Different types of vehicles (e.g. taxis, buses, online ride-hailing vehicles, and shared bikes) can be equipped with GPS receivers, which record GPS coordinates in 2-60 second intervals. The trajectory data calculated from these GPS coordinate samples can be matched to road networks and further used to derive traffic flow or speed. The advantage of using GPS trajectory data for graph-based modeling is both the low expense to collect GPS data with smartphones and the wider coverage with the massive number of vehicles, compared with traffic sensor data. However, GPS trajectory data contain no direct traffic information, which can be derived with corresponding definitions though. The data quality problems also remain with GPS trajectory data and more pre-processing steps are required, e.g., map matching.

\textit{Location-based Service Data}. GPS function is also embedded in smartphones, which can be used to collect various types of location-related data, e.g., check-in data, point-of-interest data, and route navigation application data. The pros and cons of using location-based service data are similar with GPS trajectory data. And the difference is that location-based service data are often collected in a crowd-sourced approach, with more data providers but potentially a lower data quality.

\textit{Trip Record Data}. These include departure and arrival dates/times, departure and arrival locations, and other trip information. Traffic speed and demand can derived from trip record data from various sources, e.g., taxis, ride-hailing services, buses, bikes, or even dock-less e-scooters used in~\cite{he2020dynamic}. These data can be collected in public transportation systems with mature methods, for example, by AFC (Automatic Fare Collection) in the subway and bus systems. Trip record data have the advantage of being capable of constructing multiple graph-based problems, e.g., station-level traffic flow and demand problems. They are also easier to collect in existing public transportation systems.

\textit{Traffic Report Data}. This type of data is often used for abnormal cases, e.g., anomaly report data used in~\cite{Liu2020stmfm} and traffic accident report data used in~\cite{zhou2020riskoracle, zhang2020multi1, zhou2020foresee}. Traffic report data are less used in graph-based modeling  because of their sparsity in both spatial and temporal dimensions, compared with trip record data.
    
\textit{Multimedia Data}. This type of data can be used as an additional input to deep learning models or for verifying the traffic status indicated by other data sources. Multimedia data used in the surveyed studies include the Baidu street-view images used in~\cite{qin2020graph} for traffic congestion, as well as satellite imagery data~\citep{zhang2020multi1}, and video surveillance data~\citep{shao2020estimation}. Multimedia data are also less seen in graph-based modeling because of their higher requirement for data collection, transmission and storage, compared with traffic sensor data with similar functionalities. It is also more difficult to extract precise traffic information, e.g., vehicle counts, from images or videos through image processing and object detection techniques.

\textit{Simulated Traffic Data}. In addition to observed real-world datasets, microscopic traffic simulators are also used to build virtual training and testing datasets for deep learning models. Examples in the surveyed studies include the MATES Simulator used in~\cite{fukuda2020short} and INTEGRATION software used in~\cite{ramadan2020traffic}. With many real-world datasets available, simulated traffic data are rarely used in GNN-based and more broader ML-based traffic forecasting studies. Traffic simulations have the potential of modeling unseen graphs though, e.g., evaluating a planned road topology.

\textit{Weather Data}. Traffic states are highly affected by the meteorological factors including temperature, humidity, precipitation, barometer pressure, and wind strength.

\textit{Calendar Data}. This includes the information on weekends and holidays. Because traffic patterns vary significantly between weekdays and weekends/holidays, some studies consider these two cases separately. Both weather and calendar data have been proven useful for traffic forecasting in the literature and should not be neglected in graph-based modeling as external factors.

While present road network and weather data can be easily found on the Internet, it is much more difficult to source historical traffic data, both due to data privacy concerns and the transmission and storage requirements of large data volumes. In Table~\ref{tab:open_data} we present a list of the open data resources used in the surveyed studies. Most of these open data are already cleaned or preprocessed and can be readily used for benchmarking and comparing the performance of different models in future work.

\begin{table}[!htb]
\tiny
    \centering
    \caption{Open data for traffic prediction problems.}
    \label{tab:open_data}
    \begin{tabular}{|l|p{9cm}|}
    \hline Dataset Name & Relevant Studies \\
    \hline
    METR-LA & ~\cite{li2018dcrnn_traffic, wu2019graph, xu2020ge, pan2020spatio, pan2019urban, lu2019graph, zhang2020spatio, wang2020traffic, zhang2020graphieee, boukerche2020performance, cao2020spectral, yu2019st, mengzhang2020spatial, tian2020st, chen2020dynamic, zhu2020a3t, zhang2018gaan, cirstea2019graph, shleifer2019incrementally, yang2020relational, chen2019gated, wang2020forecast, cui2020graph, zhou2020reinforced, cai2020traffic, zhou2020variational, wu2020connecting, chen2020multi, opolka2019spatio, oreshkin2020fc, jia2020dynamic, zhang2020spatial1, feng2020dynamic, xie2020istd, park2020st, song2020graph} \\
    \hline
    PeMS all & ~\cite{mallick2020graph, mallick2020transfer} \\
    \hline
    PeMS-BAY & ~\cite{li2018dcrnn_traffic, wu2019graph, zheng2020gman, pan2020spatio, pan2019urban, xu2020spatial, wang2020traffic, zhang2020graphieee, boukerche2020performance, li2020multi, cao2020spectral, xie2020sast, mengzhang2020spatial, tian2020st, shleifer2019incrementally, chen2019gated, yu20193d, wang2020forecast, cui2020graph, zhou2020reinforced, cai2020traffic, zhou2020variational, wu2020connecting, chen2020multi, oreshkin2020fc, guo2020short, zhang2020spatial1, feng2020dynamic, xie2020istd, park2020st, song2020graph} \\
    \hline
    PeMSD3 & ~\cite{song2020spatial, cao2020spectral, chen2020tssrgcn, wang2020auto, mengzhang2020spatial} \\
    \hline
    PeMSD4 & ~\cite{bai2020adaptive, huang2020lsgcn, zhang2020spatio, song2020spatial, chen2020gdcrn, tang2020general, guo2019attention, li2019hybrid, wei2020spatial, cao2020spectral, li2020traffic, yin2020multi, zhang2020attention, wang2020auto, xin2020internet, huang2020short, guo2020dynamic, mengzhang2020spatial, xu2020traffic, chen2020gst, ge2019temporal, ge2019traffic, ge2020global, zhao2020attention} \\
    \hline
    PeMSD7 & ~\cite{zhang2020novel, huang2020lsgcn, song2020spatial, xu2020spatial, tang2020general, sun2020traffic, cao2020spectral, bing2018spatio, yu2019st, chen2020tssrgcn, wang2020auto, xin2020internet, xie2020sast, mengzhang2020spatial, zhang2019spatial, ge2019temporal, ge2019traffic, ge2020global, yu20193d, zhao2020attention} \\
    \hline
    PeMSD8 & ~\cite{bai2020adaptive, huang2020lsgcn, song2020spatial, chen2020gdcrn, guo2019attention, wei2020spatial, cao2020spectral, li2020traffic, yin2020multi, zhang2020attention, wang2020auto, guo2020dynamic, mengzhang2020spatial} \\
    \hline
    Seattle Loop & ~\cite{cui2019traffic, cui2020learning, sun2020constructing, lewenfus2020joint} \\
    \hline
    T-Drive & ~\cite{pan2020spatio, pan2019urban} \\
    \hline
    SHSpeed & ~\cite{zhang2020novel, wang2018efficient, guo2019multi} \\
    \hline
    TaxiBJ & ~\cite{zhang2020spatial, wang2018graph, bai2019stg2seq} \\
    \hline
    TaxiSZ & ~\cite{zhu2020a3t, zhao2019t} \\
    \hline
    TaxiCD & ~\cite{hu2018recurrent, hu2020stochastic} \\
    \hline
    TaxiNYC & ~\cite{zhang2020spatial, sun2020predicting, zhou2019revisiting, hu2018recurrent, jin2020deep, li2020short, zheng2020spatial, xu2019incorporating, davis2020grids, du2020traffic, li2020forecaster, ye2020coupled, zhou2020foresee} \\
    \hline
    UberNYC & ~\cite{jin2020deep, ke2020joint} \\
    \hline
    DiDiChengdu & ~\cite{zhang2019graph, qu2020modeling, wang2020mtgcn, zhou2019revisiting, wang2020urban, bogaerts2020graph, li2020sdcn} \\
    \hline
    DiDiTTIChengdu & ~\cite{lu2020spatiotemporal} \\
    \hline
    DiDiXi'an & ~\cite{qu2020modeling, bogaerts2020graph} \\
    \hline
    DiDiHaikou & ~\cite{pian2020spatial, jin2020urban} \\
    \hline
    BikeDC & ~\cite{sun2020predicting, wang2020spatial} \\
    \hline
    BikeNYC & ~\cite{zhang2020spatial, sun2020predicting, wang2018graph, he2020towards, chai2018bike, lee2019demand, bai2019stg2seq, du2020traffic, ye2020coupled, wang2020spatial, guo2019bikenet, lin2018predicting} \\
    \hline 
    BikeChicago & ~\cite{chai2018bike} \\
    \hline
    SHMetro & ~\cite{liu2020physical} \\
    \hline
    HZMetro & ~\cite{liu2020physical} \\
    \hline
    \end{tabular}
\end{table}

\subsubsection{Traffic Sensor Data}
The relevant open traffic sensor data are listed as follows.

\textit{METR-LA~\footnote{Download link: \url{https://github.com/liyaguang/DCRNN}}}: This dataset contains traffic speed and volume collected from the highway of the Los Angeles County road network, with 207 loop detectors. The samples are aggregated in 5-minute intervals. The most frequently referenced time period for this dataset is from March 1st to June 30th, 2012.

\textit{Performance Measurement System (PeMS) Data~\footnote{\url{http://pems.dot.ca.gov/}}}: This dataset contains raw detector data from over 18,000 vehicle detector stations on the freeway system spanning all major metropolitan areas of California from 2001 to 2019, collected with various sensors including inductive loops, side-fire radar, and magnetometers. The samples are captured every 30 seconds and aggregated in 5-minute intervals. Each data sample contains a timestamp, station ID, district, freeway ID, direction of travel, total flow, and average speed. Different subsets of PeMS data have been used in previous studies, for example:
\begin{itemize}
    \item PeMS-BAY~\footnote{Download link: \url{https://github.com/liyaguang/DCRNN}}: This subset contains data from 325 sensors in the Bay Area from January 1st to June 30th, 2017.
    \item PeMSD3: This subset uses 358 sensors in the North Central Area. The frequently referenced time period for this dataset is September 1st to November 30th, 2018.
    \item PeMSD4: This subset uses 307 sensors in the San Francisco Bay Area. The frequently referenced time period for this dataset is January 1st to February 28th, 2018.
    \item PeMSD7: This subset uses 883 sensors in the Los Angeles Area. The frequently referenced time period for this dataset is May to June, 2012.
    \item PeMSD8: This subset uses 170 sensors in the San Bernardino Area. The frequently referenced time period for this dataset is July to August, 2016.
\end{itemize}

\textit{Seattle Loop~\footnote{Download link:~\url{https://github.com/zhiyongc/Seattle-Loop-Data}}}: This dataset was collected by inductive loop detectors deployed on four connected freeways (I-5, I-405, I-90, and SR-520) in the Seattle area, from January 1st to 31st, 2015. It contains the traffic speed data from 323 detectors. The samples are aggregated in 5-minute intervals.

\subsubsection{Taxi Data}
The open taxi datasets used in the surveyed studies are listed as follows.

\textit{T-drive~\citep{yuan2010t}}: This dataset contains a large number of taxicab trajectories collected by 30,000 taxis in Beijing from February 1st to June 2nd, 2015.

\textit{SHSpeed (Shanghai Traffic Speed)~\citep{wang2018efficient} \footnote{Download link: \url{https://github.com/xxArbiter/grnn}}}: This dataset contains 10-minute traffic speed data, derived from raw taxi trajectory data, collected from 1 to 30 April 2015, for 156 urban road segments in the central area of Shanghai, China.

\textit{TaxiBJ~\citep{zhang2017deep}}: This dataset contains inflow and outflow data derived from GPS data in more than 34,000 taxicabs in Beijing from four time intervals: (1) July 1st to October 30th, 2013; (2) March 1st to June 30th, 2014; (3) March 1st to June 30th, 2015; and (4) November 1st, 2015 to April 10th, 2016. The Beijing city map is divided into $32\times32$ grids and the time interval of the flow data is 30 minutes.

\textit{TaxiSZ~\citep{zhao2019t}~\footnote{Download link:~\url{https://github.com/lehaifeng/T-GCN}}}: This dataset is derived from taxi trajectories in Shenzhen from January 1st to 31st, 2015. It contains the traffic speed on 156 major roads of the Luohu District every 15 minutes.

\textit{TaxiCD~\footnote{~\url{https://js.dclab.run/v2/cmptDetail.html?id=175}}}: This dataset contains 1.4 billion GPS records from 14,864 taxis collected from August 3rd to 30th, 2014 in Chengdu, China. Each GPS record consists of a taxi ID, latitude, longitude, an indicator of whether the taxi is occupied, and a timestamp.

\textit{TaxiNYC\footnote{\url{http://www.nyc.gov/html/tlc/html/about/trip_record_data.shtml}}}: The taxi trip records in New York starting from 2009, in both yellow and green taxis. Each trip record contains pick-up and drop-off dates/times, pick-up and drop-off locations, trip distances, itemized fares, rate types, payment types, and driver-reported passenger counts.

\subsubsection{Ride-hailing Data}
The open ride-hailing data used in the surveyed studies are listed as follows.

\textit{UberNYC~\footnote{\url{https://github.com/fivethirtyeight/uber-tlc-foil-response}}}: This dataset comes from Uber, which is one of the largest online ride-hailing companies in the USA, and is provided by the NYC Taxi \& Limousine Commission (TLC). It contains data from over 4.5 million Uber pickups in New York City from April to September 2014, and 14.3 million more Uber pickups from January to June 2015.

\textit{Didi GAIA Open Data~\footnote{\url{https://outreach.didichuxing.com/research/opendata/}}}: This open data plan is supported by Didi Chuxing, which is one of the largest online ride-hailing companies in China.
\begin{itemize}
    \item DiDiChengdu: This dataset contains the trajectories of DiDi Express and DiDi Premier drivers within Chengdu, China. The data contains trips from October to November 2016.
    \item DiDiTTIChengdu: This dataset represents the DiDi Travel Time Index Data in Chengdu, China in the year of 2018, which contains the average speed of major roads every 10 minutes.
    \item DiDiXi'an: This dataset contains the trajectories of DiDi Express and DiDi Premier drivers within Xi'an, China. The data contains trips from October to November 2016.
    \item DiDiHaikou: The dataset contains DiDi Express and DiDi Premier orders from May 1st to October 31st, 2017 in the city of Haikou, China, including the coordinates of origins and destinations, pickup and drop-off timestamps, as well as other information.
\end{itemize}

\subsubsection{Bike Data}
The open bike data used in the surveyed studies are listed as follows.

\textit{BikeNYC~\footnote{~\url{https://www.citibikenyc.com/system-data}}}: This dataset is from the NYC Bike System, which contains 416 stations. The frequently referenced time period for this dataset is from 1st July, 2013 to 31th December, 2016.

\textit{BikeDC~\footnote{~\url{https://www.capitalbikeshare.com/system-data}}}: This dataset is from the Washington D.C. Bike System, which contains 472 stations. Each record contains trip duration, start and end station IDs, and start and end times.

\textit{BikeChicago~\footnote{\url{https://www.divvybikes.com/system-data}}}: This dataset is from the Divvy System Data in Chicago, from 2015 to 2020.

\subsubsection{Subway Data}
The subway data referenced in the surveyed studies are listed as follows.

\textit{SHMetro~\citep{liu2020physical}~\footnote{Download link:~\url{https://github.com/ivechan/PVCGN}}}: This dataset is derived from 811.8 million transaction records of the Shanghai metro system collected from July 1st to September 30th, 2016. It contains 288 metro stations and 958 physical edges. The inflow and outflow of each station are provided in 15 minute intervals.

\textit{HZMetro~\citep{liu2020physical}~\footnote{Download link:~\url{https://github.com/ivechan/PVCGN}}}: This dataset is similar to SHMetro, from the metro system in Hangzhou, China, in January 2019. It contains 80 metro stations and 248 physical edges, and the aggregation time length is also 15 minutes.

\subsection{Open Source Codes}
Several open source frameworks for implementing general deep learning models, most of which are built with the Python programming language, can be accessed online, e.g. TensorFlow~\footnote{\url{https://www.tensorflow.org/}}, Keras~\footnote{\url{https://keras.io/}}, PyTorch~\footnote{\url{https://pytorch.org/}}, and MXNet~\footnote{\url{https://mxnet.apache.org/}}. Additional Python libraries designed for implementing GNNs are available. These include DGL~\footnote{\url{https://www.dgl.ai/}}, pytorch\_geometric~\footnote{\url{https://pytorch-geometric.readthedocs.io/}}, and Graph Nets~\footnote{\url{https://github.com/deepmind/graph_nets}}.

Many authors have also released open-source implementations of their proposed models. The open source projects for traffic flow, traffic speed, traffic demand, and other problems are summarized in Tables~\ref{tab:code_traffic_flow}, ~\ref{tab:code_traffic_speed}, ~\ref{tab:code_traffic_demand}, and~\ref{tab:code_other_problems}, respectively.  In these open source projects, TensorFlow and PyTorch are the two frameworks that are used most frequently.

\begin{table*}[!htb]
\tiny
    \centering
    \caption{Open source projects for traffic flow related problems.}
    \label{tab:code_traffic_flow}
    \begin{tabular}{|l|l|l|p{2.5cm}|p{5cm}|}
        \hline Article & Year & Framework & Problem & Link \\ 
        \hline
        ~\cite{zheng2020gman} & 2020 & TensorFlow & Road Traffic Flow, Road Traffic Speed & \url{https://github.com/zhengchuanpan/GMAN} \\
        \hline
        ~\cite{bai2020adaptive} & 2020 & PyTorch & Road Traffic Flow & \url{https://github.com/LeiBAI/AGCRN} \\
        \hline
        ~\cite{song2020spatial} & 2020 & MXNet & Road Traffic Flow & \url{https://github.com/wanhuaiyu/STSGCN} \\
        \hline 
        ~\cite{tang2020general} & 2020 & TensorFlow & Road Traffic Flow & \url{https://github.com/sam101340/GAGCN-BC-20200720} \\
        \hline
        ~\cite{wang2020auto} & 2020 & MXNet, PyTorch & Road Traffic Flow & \url{https://github.com/zkx741481546/Auto-STGCN} \\
        \hline
        ~\cite{guo2020dynamic} & 2020 & PyTorch & Road Traffic Flow, Road Traffic Speed & \url{https://github.com/guokan987/DGCN} \\
        \hline
        ~\cite{mengzhang2020spatial} & 2020 & MXNet & Road Traffic Flow, Road Traffic Speed & \url{https://github.com/MengzhangLI/STFGNN} \\
        \hline
        ~\cite{tian2020st} & 2020 & PyTorch, DGL & Road Traffic Flow & \url{https://github.com/Kelang-Tian/ST-MGAT} \\
        \hline
        ~\cite{xiong2020dynamic} & 2020 & TensorFlow & Road OD Flow & \url{https://github.com/alzmxx/OD_Prediction} \\
        \hline
        ~\cite{peng2020spatial} & 2020 & Keras & Road Station-level Subway Passenger Flow, Station-level Bus Passenger Flow, Regional Taxi Flow & \url{https://github.com/RingBDStack/GCNN-In-Traffic} \\
        \hline
        ~\cite{qiu2020topological} & 2020 & Pytorch & Regional Taxi Flow & \url{https://github.com/Stanislas0/ToGCN-V2X} \\
        \hline
        ~\cite{yeghikyan2020learning} & 2020 & PyTorch & Regional OD Taxi Flow & \url{https://github.com/FelixOpolka/Mobility-Flows-Neural-Networks} \\
        \hline
        ~\cite{zhang2020deep} & 2020 & Keras & Station-level Subway Passenger Flow & \url{https://github.com/JinleiZhangBJTU/ResNet-LSTM-GCN} \\
        \hline
        ~\cite{zhang2020multi} & 2020 & Keras & Station-level Subway Passenger Flow & \url{https://github.com/JinleiZhangBJTU/Conv-GCN} \\
        \hline
        ~\cite{liu2020physical} & 2020 & PyTorch & Station-level Subway Passenger Flow & \url{https://github.com/ivechan/PVCGN} \\
        \hline
        ~\cite{ye2020multi} & 2020 & Keras & Station-level Subway Passenger Flow & \url{https://github.com/start2020/Multi-STGCnet} \\
        \hline
        ~\cite{pan2019urban} & 2019 & MXNet, DGL & Road Traffic Flow, Road Traffic Speed & \url{https://github.com/panzheyi/ST-MetaNet} \\
        \hline
        ~\cite{guo2019attention} & 2019 & MXNet & Road Traffic Flow & \url{https://github.com/wanhuaiyu/ASTGCN} \\
        \hline
        ~\cite{guo2019attention} & 2019 & PyTorch & Road Traffic Flow & \url{https://github.com/wanhuaiyu/ASTGCN-r-pytorch} \\
        \hline
        ~\cite{wang2018efficient} & 2018 & PyTorch & Road Traffic Flow & \url{https://github.com/xxArbiter/grnn} \\
        \hline
        ~\cite{bing2018spatio} & 2018 & TensorFlow & Road Traffic Flow & \url{https://github.com/VeritasYin/STGCN_IJCAI-18} \\
        \hline
        ~\cite{li2018graph} & 2018 & Keras & Station-level Subway Passenger Flow & \url{https://github.com/RingBDStack/GCNN-In-Traffic} \\
        \hline
        ~\cite{chai2018bike} & 2018 & TensorFlow & Bike Flow & \url{https://github.com/Di-Chai/GraphCNN-Bike} \\
        \hline
    \end{tabular}
\end{table*}

\begin{table*}[!htb]
\tiny
    \centering
    \caption{Open source projects for traffic speed related problems.}
    \label{tab:code_traffic_speed}
    \begin{tabular}{|p{2cm}|l|l|p{2.5cm}|p{5cm}|}
    \hline Article & Year & Framework & Problem & Link \\
    \hline
    ~\cite{zhang2020spatial} & 2020 & Keras & Road Traffic Speed & \url{https://github.com/jillbetty001/ST-CGA} \\
    \hline
    ~\cite{zhu2020a3t} & 2020 & TensorFlow & Road Traffic Speed & \url{https://github.com/lehaifeng/T-GCN/tree/master/A3T} \\
    \hline
    ~\cite{yang2020relational} & 2020 & TensorFlow & Road Traffic Speed & \url{https://github.com/fanyang01/relational-ssm} \\
    \hline
    ~\cite{wu2020connecting} & 2020 & PyTorch & Road Traffic Speed & \url{https://github.com/nnzhan/MTGNN} \\
    \hline
    ~\cite{mallick2020transfer} & 2020 & TensorFlow & Road Traffic Speed & \url{https://github.com/tanwimallick/TL-DCRNN} \\
    \hline
    ~\cite{chen2020graph} & 2020 & PyTorch & Road Traffic Speed & \url{https://github.com/Fanglanc/DKFN} \\
    \hline
    ~\cite{lu2020spatiotemporal} & 2020 & PyTorch & Road Traffic Speed & \url{https://github.com/RobinLu1209/STAG-GCN} \\
    \hline
    ~\cite{guopeng2020dynamic} & 2020 & TensorFlow, Keras & Road Traffic Speed & \url{https://github.com/RomainLITUD/DGCN_traffic_forecasting} \\
    \hline
    ~\cite{shen2020ttpnet} & 2020 & PyTorch & Road Travel Time & \url{https://github.com/YibinShen/TTPNet} \\
    \hline
    ~\cite{hong2020heteta} & 2020 & TensorFlow & Time of Arrival & \url{https://github.com/didi/heteta} \\
    \hline
    ~\cite{wu2019graph} & 2019 & PyTorch & Road Traffic Speed & \url{https://github.com/nnzhan/Graph-WaveNet} \\
    \hline
    ~\cite{shleifer2019incrementally} & 2019 & PyTorch & Road Traffic Speed & \url{https://github.com/sshleifer/Graph-WaveNet} \\
    \hline
    ~\cite{zhao2019t} & 2019 & TensorFlow & Road Traffic Speed & \url{https://github.com/lehaifeng/T-GCN} \\
    \hline
    ~\cite{cui2019traffic} & 2019 & TensorFlow & Road Traffic Speed & \url{https://github.com/zhiyongc/Graph_Convolutional_LSTM} \\
    \hline
    ~\cite{jepsen2019graph, jepsen2020relational} & 2019 & MXNet & Road Traffic Speed & \url{https://github.com/TobiasSkovgaardJepsen/relational-fusion-networks} \\
    \hline
    ~\cite{li2018dcrnn_traffic} & 2018 & TensorFlow & Road Traffic Speed & \url{https://github.com/liyaguang/DCRNN} \\
    \hline
    ~\cite{li2018dcrnn_traffic} & 2018 & PyTorch & Road Traffic Speed & \url{https://github.com/chnsh/DCRNN_PyTorch} \\
    \hline
    ~\cite{zhang2018gaan} & 2018 & MXNet & Road Traffic Speed & \url{https://github.com/jennyzhang0215/GaAN} \\
    \hline
    ~\cite{liao2018deep} & 2018 & TensorFlow & Road Traffic Speed & \url{https://github.com/JingqingZ/BaiduTraffic} \\
    \hline
    ~\cite{mohanty2018graph, mohanty2020region} & 2018 & TensorFlow & Traffic Congestion & \url{https://github.com/sudatta0993/Dynamic-Congestion-Prediction} \\
    \hline
    \end{tabular}
\end{table*}

\begin{table*}[!htb]
\tiny
    \centering
    \caption{Open source projects for traffic demand related problems.}
    \label{tab:code_traffic_demand}
    \begin{tabular}{|l|l|l|p{2.5cm}|p{5cm}|}
    \hline Article & Year & Framework & Problem & Link \\
    \hline
    ~\cite{hu2020stochastic} & 2020 & TensorFlow & Taxi Demand & \url{https://github.com/hujilin1229/od-pred} \\
    \hline
    ~\cite{davis2020grids} & 2020 & TensorFlow, PyTorch & Taxi Demand & \url{https://github.com/NDavisK/Grids-versus-Graphs} \\
    \hline
    ~\cite{ye2020coupled} & 2020 & PyTorch & Taxi Demand, Bike Demand & \url{https://github.com/Essaim/CGCDemandPrediction} \\
    \hline
    ~\cite{lee2019demand} & 2019 & TensorFlow, Keras & Ride-hailing Demand, Bike Demand, Taxi Demand & \url{https://github.com/LeeDoYup/TGGNet-keras} \\
    \hline
    ~\cite{ke2019predicting} & 2019 & Keras & Taxi Demand & \url{https://github.com/kejintao/ST-ED-RMGC} \\
    \hline
    \end{tabular}
\end{table*}

\begin{table*}[!htb]
\tiny
    \centering
    \caption{Open source projects for other problems.}
    \label{tab:code_other_problems}
    \begin{tabular}{|l|l|l|p{2.5cm}|p{5cm}|}
    \hline Article & Year & Framework & Problem & Link \\
    \hline
    ~\cite{zhou2020riskoracle} & 2020 & TensorFlow & Traffic Accident & \url{https://github.com/zzyy0929/AAAI2020-RiskOracle/} \\
    \hline
    ~\cite{yu2020deep} & 2020 & PyTorch, DGL & Traffic Accident & \url{https://github.com/yule-BUAA/DSTGCN} \\
    \hline
    ~\cite{zhang2020semi} & 2020 & PyTorch, DGL & Parking Availability & \url{https://github.com/Vvrep/SHARE-parking_availability_prediction-Pytorch} \\
    \hline
    ~\cite{wang2020evaluation} & 2020 & TensorFlow & Transportation Resilience & \url{https://github.com/Charles117/resilience_shenzhen} \\
    \hline
    ~\cite{wright2019neural} & 2019 & TensorFlow, Keras & Lane Occupancy & \url{https://github.com/mawright/trafficgraphnn} \\
    \hline
    \end{tabular}
\end{table*}

\subsection{State-of-the-art Performance}
\label{sec:benchmarking}
It is known that different works use different datasets and it is very hard to assess the relative performance of different state-of-the-art models~\citep{tedjopurnomo2020survey}. Even for those studies using the same dataset, different subsets may be used. Different preprocessing techniques, e.g., the missing data imputation method, and different evaluation settings, e.g., the training/validation/test subset split ratio, also cause incomparable results. Considering these difficulties, we only summarize those comparable results for the most frequently used datasets from the surveyed studies in this part.

Some commonly used evaluation metrics, namely, RMSE, MAE and MAPE, are defined as follows:
\begin{itemize}
\item $\text{RMSE}(\mathbf{y}, \mathbf{\hat{y}})=\sqrt{\frac{1}{M}\sum_{i=1}^{M}{(y_i - \hat{y}_i)^2}}$;
\item $\text{MAE}(\mathbf{y}, \mathbf{\hat{y}})=\frac{1}{M}\sum_{i=1}^{M}|y_i - \hat{y}_i|$;
\item $\text{MAPE}(\mathbf{y}, \mathbf{\hat{y}})=\frac{1}{M}\sum_{i=1}^{M}\frac{|y_i - \hat{y}_i|}{y_i}$;
\end{itemize}
\noindent where $\mathbf{y}$ denotes the true values, $\mathbf{\hat{y}}$ denotes the predicted values, and $M$ is the number of values to predict. A lower RMSE or MAE value indicates a better prediction performance. The summary for the state-of-the-art performance is shown in Table~\ref{tab:performance}, with all or some of the above evaluation metrics and best values in bold. The default prediction time period is 60 minutes in Table~\ref{tab:performance} unless otherwise specified. Some classical baselines are also listed for comparison if available, e.g., DCRNN~\citep{li2018dcrnn_traffic}, STGCN~\citep{bing2018spatio} and Graph WaveNet~\citep{wu2019graph}. Interested readers are recommended to check the experimental details in relevant studies.

\begin{table}[!htb]
\tiny
    \centering
    \caption{State-of-the-art performance for traffic prediction problems.}
    \label{tab:performance}
    \begin{tabular}{|l|l|l|l|l|}
    \hline
    Dataset & RMSE & MAE & MAPE & Relevant Studies \\
    \hline
    \multirow{6}{*}{METR-LA} & 7.59 & 3.60 & 10.5\% & DCRNN \\
    \cline{2-5}
    & 7.40 & 3.55 & 10.0\% & ST-UNet~\citep{yu2019st} \\
    \cline{2-5}
    & 7.37 & 3.53 & 10.0\% & Graph WaveNet \\
    \cline{2-5}
    & 7.20 & 3.30 & 9.7\% & SLCNN~\citep{zhang2020spatio} \\
    \cline{2-5}
    & 6.68 & 3.28 & 9.08\% & Traffic Transformer~\citep{cai2020traffic} \\
    \cline{2-5}
    & \textbf{6.40} & \textbf{3.18} & \textbf{8.81\%} & STFGNN~\citep{mengzhang2020spatial} \\
    \hline
    \multirow{6}{*}{PeMS-BAY} & 4.74 & 2.07 & 4.9\% & DCRNN \\
    \cline{2-5}
    & 4.53 & 2.03 & 4.8\% & SLCNN~\citep{zhang2020spatio} \\
    \cline{2-5}
    & 4.52 & 1.95 & 4.63\% & Graph WaveNet \\
    \cline{2-5}
    & 4.32 & 1.86 & 4.31\% & GMAN~\citep{zheng2020gman} \\
    \cline{2-5}
    & 4.36 & 1.77 & 4.29\% & Traffic Transformer~\citep{cai2020traffic} \\
    \cline{2-5}
    & \textbf{3.74} & \textbf{1.66} & \textbf{3.77\%} & STFGNN~\citep{mengzhang2020spatial} \\
    \hline
    \multirow{4}{*}{PeMSD3}  & 30.31 & 18.18 & 18.91\% & DCRNN \\
    \cline{2-5}
    & 30.12 & 17.49 & 17.15\% & STGCN \\
    \cline{2-5}
    & 32.94 & 17.48 & 16.78\% & Graph WaveNet \\
    \cline{2-5}
    & \textbf{28.34} & \textbf{16.77} & \textbf{16.30\%} & STFGNN~\citep{mengzhang2020spatial} \\
    \hline
    \multirow{5}{*}{PeMSD4} & 39.70 & 25.45 & 17.29\% & Graph WaveNet \\
    \cline{2-5}
    & 34.89 & 21.16 & 13.83\% & STGCN \\
    \cline{2-5}
    & 33.44 & 21.22 & 14.17\% & DCRNN \\
    \cline{2-5}
    & 32.26 & \textbf{19.83} & \textbf{12.97\%} & AGCRN~\citep{bai2020adaptive} \\
    \cline{2-5}
    & \textbf{31.88} & \textbf{19.83} & 13.02\% & STFGNN~\citep{mengzhang2020spatial} \\
    \hline
    \multirow{4}{*}{PeMSD7} & 42.78 & 26.85 & 12.12\% & Graph WaveNet \\
    & 38.78 & 25.38 & 11.08\% & STGCN \\
    \cline{2-5}
    & 38.58 & 25.30 & 11.66\% & DCRNN \\
    \cline{2-5}
    & \textbf{35.80} & \textbf{22.07} & \textbf{9.21\%} & STFGNN~\citep{mengzhang2020spatial} \\
    \hline
    \multirow{5}{*}{PeMSD8} & 31.05 & 19.13 & 12.68\% & Graph WaveNet \\
    \cline{2-5}
    & 27.09 & 17.50 & 11.29\% & STGCN \\
    \cline{2-5}
    & 26.36 & 16.82 & 10.92\% & DCRNN \\
    \cline{2-5}
    & 26.22 & 16.64 & 10.60\% & STFGNN~\citep{mengzhang2020spatial} \\
    \cline{2-5}
    & \textbf{25.22} & \textbf{15.95} & \textbf{10.09\%} & AGCRN~\citep{bai2020adaptive} \\
    \hline
    \multirow{2}{*}{Seattle Loop} & 8.22 & 4.64 & 11.18\% & DCRNN \\
    \cline{2-5}
    & \textbf{3.59} & \textbf{2.45} & \textbf{5.90\%} & GLT-GCRNN~\citep{sun2020constructing} \\
    \hline
    \multirow{6}{*}{TaxiSZ} & 4.76 & 3.38 & N/A & Graph WaveNet \\
    \cline{2-5}
    & 4.64 & 3.31 & N/A & DCRNN \\
    \cline{2-5}
    & 4.13 & 2.79 & N/A & T-GCN~\citep{zhao2019t} \\
    \cline{2-5}
    & 4.13 & 2.76 & N/A & STGCN \\
    \cline{2-5}
    & 4.10 & 2.77 & N/A & AST-GCN~\citep{zhu2020ast} \\
    \cline{2-5}
    & \textbf{3.97} & \textbf{2.74} & N/A & A3T-GCN~\citep{zhu2020a3t} \\
    \hline
    \multirow{4}{*}{TaxiNYC (30 min)} & 22.65 & 18.46 & N/A & STGCN \\
    \cline{2-5}
    & 14.79 & 8.43 & N/A & DCRNN \\
    \cline{2-5}
    & 13.07 & 8.10 & N/A & Graph WaveNet \\
    \cline{2-5}
    & \textbf{9.56} & \textbf{5.50} & N/A & CCRNN~\citep{ye2020coupled} \\
    \hline
    \multirow{4}{*}{BikeNYC (30 min)} & 3.60 & 2.76 & N/A & STGCN \\
    \cline{2-5}
    & 3.29 & 1.99 & N/A & Graph WaveNet \\
    \cline{2-5}
    & 3.21 & 1.90 & N/A & DCRNN \\
    \cline{2-5}
    & \textbf{2.84} & \textbf{1.74} & N/A & CCRNN~\citep{ye2020coupled} \\
    \hline
    \end{tabular}
\end{table}

Since the relevant studies of applying GNNs for traffic forecasting are growing everyday, the results listed in this part are not guaranteed to be the latest ones and the readers are recommended to follow our Github repository to track latest results.

\section{Challenges and Future Directions}
\label{sec:challenges}
In this section, we discuss general challenges for traffic prediction problems as well as specific new challenges when GNNs are involved. While GNNs achieve a better forecasting performance, they are not the panacea. Some existing challenges from the border topic of traffic forecasting remain unsolved in current graph-based studies. Based on these challenges, we discuss possible future directions as well as early attempts in these directions. Some of these future directions are inspired from the border traffic forecasting research and remain insightful for the graph-based modeling approach. We would also highlight the special opportunities with GNNs.

\subsection{Challenges}
\subsubsection{Heterogeneous Data}
Traffic prediction problems involve both spatiotemporal data and external factors, e.g., weather and calendar information. Heterogeneous data fusion is a challenge that is not limited to the traffic domain. GNNs have enabled significant progress by taking the underlying graph structures into consideration. However, some challenges remain; for example, geographically close nodes may not be the most influential, both for CNN-based and GNN-based approaches. Another special challenge for GNNs is that the underlying graph information may not be correct or up to date. For example, the road topology data of OpenStreetMap, an online map services, are collected in a crowd-sourced approach, which may be inaccurate or lagged behind the real road network. The spatial dependency relationship extracted by GNNs with these inaccurate data may decrease the forecasting accuracy.

Data quality concerns present an additional challenge with problems such as missing data, sparse data and noise potentially compromising forecasting results. Most of the surveyed models are only evaluated with processed high-quality datasets. A few studies do, however, take data quality related problems into consideration, e.g., using the Kalman filter to deal with the sensor data bias and noise~\citep{chen2020graph}, infilling missing data with moving average filters~\citep{hasanzadeh2019piecewise} or linear interpolation~\citep{agafonov2020traffic, chen2020graph}. Missing data problem could be more common in GNNs, with the potential missing phenomena happening with historical traffic data or underlying graph information, e.g., GCNs are proposed to fill data gaps in missing OD flow problems~\citep{yao2020spatial}.

Traffic anomalies (e.g., congestion) are an important external factor that may affect prediction accuracy and it has been proven that under congested traffic conditions a deep neural network may not perform as well as under normal traffic conditions~\citep{mena2020comprehensive}. However, it remains a challenge to collect enough anomaly data to train deep learning models (including GNNs) in both normal and anomalous situations. The same concern applies for social events, public holidays, etc.

Challenges also exist for data privacy in the transportation domain. As discussed in Section~\ref{sec:data}, many open data are collected from individual mobile devices in a crowd sourcing approach. The data administrator must guarantee the privacy of individuals who contribute their personal traffic data, as the basis for encouraging a further contribution. Different techniques may be used, e.g., privacy-preserving data publishing techniques and privacy-aware data structures without personal identities.

\subsubsection{Multi-task Performance}
For the public service operation of ITSs, a multi-task framework is necessary to incorporate all the traffic information and predict the demand of multiple transportation modes simultaneously. For example, knowledge adaption is proposed to adapt the relevant knowledge from an information-intensive source to information-sparse sources for demand prediction~\citep{li2020knowledge}. Related challenges lie in data format incompatibilities as well as the inherent differences in spatial or temporal patterns. While some of the surveyed models can be used for multiple tasks, e.g., traffic flow and traffic speed prediction on the same road segment, most can only be trained for a single task at one time. 

Multi-task forecasting is a bigger challenge in graph-based modeling because different tasks may use different graph structures, e.g., road-level and station-level problems use different graphs and thus are difficult to be solved with a single GNN model. Some efforts that have been made in GNN-based models for multi-task prediction include taxi departure flow and arrival flow~\citep{chen2020multitask}, region-flow and transition-flow~\citep{wang2020mtgcn}, crowd flows, and OD of the flows~\citep{wang2020multi}. However, most of the existing attempts are based on the same graph with multiple outputs generated by feed forward layers. Nonetheless, GNN-based multi-task prediction for different types of traffic forecasting problems is a research direction requiring significant further development, especially those requiring multiple graph structures.

\subsubsection{Practical Implementation}
A number of challenges prevent the practical implementation of the models developed in the surveyed studies in city-scale ITSs.

First, there is significant bias introduced by the small amount of data considered in the existing GNN-based studies which, in most cases, spans less than one year. The proposed solutions are therefore not necessarily applicable to different time periods or different places. If longer traffic data are to be used in GNNs, the corresponding change of the underlying traffic infrastructures should be recorded and updated, which increases both the expense and difficulty of the associated data collection process in practice.

A second challenge is the computation scalability of GNNs. To avoid the huge computation requirements of the large-scale real-world traffic network graphs, only a subset of the nodes and edges are typically considered. For example, most studies only use a subset of the PeMS dataset when considering the road traffic flow or speed problems. Their results can therefore only be applied to the selected subsets. Graph partitioning and parallel computing infrastructures have been proposed for solving this problem. The traffic speed and flow of the entire PeMS dataset with 11,160 traffic sensor locations are predicted simultaneously in~\cite{mallick2020graph}, using a graph-partitioning method that decomposes a large highway network into smaller networks and trains a single DCRNN model on a cluster with graphics processing units (GPUs). However, increased modeling power can only improve the state-of-the-art results with narrow performance margins, compared to statistical and machine learning models with less complex structures and computational requirements.

A third challenge is presented by changes in the transportation networks and infrastructure, which are essential to build the graphs in GNNs. The real-world network graphs change when road segments or bus lines are added or removed. Points-of-interest in a city also change when new facilities are built. Static graph formulations are not enough for handling these situations. Some efforts have been made to solve this problem with promising results. For example, a dynamic Laplacian matrix estimator is proposed to find the change of Laplacian matrix, according to changes in spatial dependencies hidden in the traffic data~\citep{diao2019dynamic}, and a Data Adaptive Graph Generation (DAGG) module is proposed to infer the inter-dependencies between different traffic series automatically, without using pre-defined graphs based on spatial connections~\citep{bai2020adaptive}.

\subsubsection{Model Interpretation}
The challenge of model interpretation is a point of criticism for all ``black-box’’ machine learning or deep learning models, and traffic forecasting tasks are no exception~\citep{wu2018hybrid, barredo2019lies}. While there have been remarkable progresses for visualizing and explaining other deep neural network structures, e.g., CNNs, the development of post-processing techniques to explain the predictions made by GNNs is still in an early phase~\citep{baldassarre2019explainability, pope2019explainability, ying2019gnnexplainer} and the application of these techniques to the traffic forecasting domain has not yet been addressed.

Compared with other similar forecasting problems in other domains, lack of model interpretation may be a more severe problem in the transportation domain, when complex data types and representations of heterogeneous traffic data make it more challenging to design an interpretable deep learning model, compared with other data formats, e.g., images and text. While some efforts have been made to incorporate the state space model to increase the model interpretation for traffic forecasting~\citep{li2019learning}, this problem has not fully solved, especially for GNN-based models.

\subsection{Future Directions}
\subsubsection{Centralized Data Repository}
A centralized data repository for GNN-based traffic forecasting resources would facilitate objective comparison of the performance of different models and be an invaluable contribution to the field. This future direction is proposed for the challenge of heterogeneous data as well as the data quality problem. Another unique feature of this repository could be the inclusion of graph-related data, which have not be provided directly in previous traffic forecasting studies.

Some criteria for building such data repositories, e.g. a unified data format, tracking of dataset versions, public code and ranked results, and sufficient record lengths (longer than a year ideally), have been discussed in previous surveys~\citep{manibardo2020deep}. Compiling a centralized and standardized data repository is particularly challenging for GNN-based models where natural graphs are collected and stored in a variety of data formats (e.g. Esri Shapefile and OSM XML used by Openstreetmap are used for digital maps in the GIS community) and various different similarity graphs can be constructed from the same traffic data in different models.

Some previous attempts in this direction have been made in the machine learning community, e.g. setting benchmarks for several traffic prediction tasks in Papers With Code~\footnote{\url{https://paperswithcode.com/task/traffic-prediction}}, and in data science competitions, e.g., the Traffic4cast competition series~\footnote{\url{https://www.iarai.ac.at/traffic4cast/}}. However, the realization of a centralized data repository remains an open challenge.

A centralized data repository is also the basis for benchmarking traffic prediction, which is previously discussed in Section~\ref{sec:benchmarking}. With more and more GNN-based models being proposed, it becomes even more difficult to compare different models and validate the effectiveness of new traffic forecasting methods without a considerable effort, when a standardized benchmark dataset and consistent experimental settings have not been established yet. The most close one is the PeMS dataset, but it covers the road-level case only and more efforts are still needed, especially for the remaining cases.

\subsubsection{Traffic Graph Design}
While various graphs have been constructed in the surveyed studies as discussed in Section~\ref{sec:graphs} and have been proven successful to some extent, most of them are natural graphs based on a real-world transportation system, e.g. the road network or subway system, as the current development status. And most of the graphs used are static, instead of dynamic ones. One specific direction that is not fully considered before is the design of transportation knowledge graph. As an important tool for knowledge integration, knowledge graph is a complex relational network that consists of concepts, entities, entity relations and attributes~\citep{yin2020comprehensive}. The transportation knowledge graph helps to leverage the traffic semantic information to improve the forecasting performance. And the challenge is to extract the hidden transportation domain knowledge from multi-source and heterogeneous traffic data.

\subsubsection{Combination with Other Techniques}
GNNs may be combined with other advanced techniques to overcome some of their inherent challenges and achieve better performance.

\textit{Data Augmentation}. Data augmentation has been proven effective for boosting the performance of deep learning models, e.g. in image classification tasks and time series prediction tasks. Data augmentation is proposed for the challenge of the possible forecasting bias introduced by the small amount of available data. However, due to the complex structure of graphs, it is more challenging to apply data augmentation techniques to GNNs. Recently, data augmentation for GNNs has proven helpful in semi-supervised node classification tasks~\citep{zhao2021data}. However, it remains a question whether data augmentation may be effective in traffic forecasting GNN applications.

\textit{Transfer Learning}. Transfer learning utilizes knowledge or models trained for one task to solve related tasks, especially those with limited data. In the image classification field, pre-trained deep learning models from the ImageNet or MS COCO datasets are widely used in other problems. In traffic prediction problems, where a lack of historical data is a frequent problem, transfer learning is a possible solution. For GNNs, transfer learning can be used from a graph with more historical traffic data for the model training process to another graph with less available data. Transfer learning can also be used for the challenge caused by the changes in the transportation networks and infrastructure, when new stations or regions have not accumulated enough historical traffic data to train a GNN model. A novel transfer learning approach for DCRNN is proposed in~\cite{mallick2020transfer}, so that a model trained on data-rich regions of highway network can be used to predict traffic on unseen regions of the highway network. The authors demonstrated the efficacy of model transferability between the San Francisco and Los Angeles regions using different parts of the California road network from the PeMS.

\textit{Meta-learning}. Meta-learning, or learning how to learn, has recently become a potential learning paradigm that can absorb information from a task and effectively generalize it to an unseen task. Meta-learning is proposed for the challenge of GNN-based multi-task prediction, especially those involving mutiple graphs. There are different types of meta learning methods and some of them are combined with graph structures for describing relationships between tasks or data samples~\citep{garcia2017few, liu2019learning}. Based on a deep meta learning method called network weight generation, ST-MetaNet$^+$ is proposed in~\cite{pan2020spatio}, which leverages the meta knowledge extracted from geo-graph attributes and dynamic traffic context learned from traffic states to generate the parameter weights in graph attention networks and RNNs, so that the inherent relationships between diverse types of spatiotemporal correlations and geo-graph attributes can be captured.

\textit{Generative Adversarial Network (GAN)~\citep{goodfellow2014generative}}. GAN is a machine learning framework that has two components, namely, a generator, which learns to generate plausible data, and a discriminator, which learns to distinguish the generator's fake data from real data. After training to a state of Nash equilibrium, the generator may generate undistinguished data, which helps to expand the training data size for many problems, including GNN-based traffic forecasting. GAN is proposed for the challenges caused by the small data amount used in previous studies or the changes in the transportation networks and infrastructure when not enough historical traffic data are available. In~\cite{xu2020ge}, the road network is used directly as the graph, in which the nodes are road state detectors and the edges are built based on their adjacent links. DeepWalk is used to embed the graph and the road traffic state sensor information is transferred into a low-dimensional space. Then, the Wasserstein GAN (WGAN)~\citep{arjovsky2017wasserstein} is used to train the traffic state data distribution and generate predicted results. Both public traffic flow (i.e., Caltrans PeMSD7) and traffic speed (i.e., METR-LA) datasets are used for evaluation, and the results demonstrate the effectiveness of the GAN-based solution when used in graph-based modeling.

\textit{Automated Machine Learning (AutoML)}. The application of machine learning requires considerable manual intervention in various aspects of the process, including feature extraction, model selection, and parameter adjustment. AutoML automatically learns the important steps related to features, models, optimization, and evaluation, so that machine learning models can be applied without manual intervention. AutoML would help to improve the implementation of machine learning models, including GNNs. AutoML is proposed for the challenge for computational requirements in graph-based modeling, in which case the hyper parameter tuning for GNNs can be more efficient with state-of-the-art AutoML techniques. An early attempt to combine AutoML with GNNs for traffic prediction problems is an Auto-STGCN algorithm, proposed in~\cite{wang2020auto}. This algorithm searches the parameter space for STGCN models quickly based on reinforcement learning and generates optimal models automatically for specific scenarios.

\textit{Bayesian Network}. Most of the existing studies aim for deterministic models that make mean predictions. However, some traffic applications rely on uncertainty estimates for the future situations. To tackle this gap, the Bayesian network, which is a type of probabilistic graphical model using Bayesian inference for probability computations, is a promising solution. The combination of GNNs with Bayesian networks is proposed for the challenge of GNN model interpretation. With probabilistic predictions, uncertainty estimates are generated for the future situations, especially the chance of extreme traffic states. A similar alternative is Quantile Regression, which estimates the quantile function of a distribution at chosen points, combined with Graph WaveNet for uncertainty estimates~\citep{maas2020uncertainty}. 

\subsubsection{Applications in Real-World ITS Systems}
Last but not the least, most of the surveyed GNN-based studies are only based on the simulations with historical traffic data, without being validated or deployed in real-world ITS systems. However, there are a number of potential applications, especially for GNN-based models with the better forecasting performance. To name a few potential cases, the GNN-based forecasting model can be used for traffic light control in signalized intersections, when each intersection is modeled as a node in the graph and the corresponding traffic flow forecasting result can be used to design the traffic light control strategy. Another example is the application in map service and navigation applications, in which each road segment is modeled as a node in the graph and the corresponding traffic speed and travel time forecasting result can be used to calculate the estimated time of arrival. A third example is the application in online ride-hailing service providers, e.g., Uber and Lyft, in which each region is modeled as a node and the corresponding ride-hailing demand forecasting can be used to design a more profitable vehicle dispatching and scheduling system. Inspired by these potential application scenarios, there are a lot of potential research opportunities for researchers from both the academia and the industry.

\section{Conclusion}
\label{sec:conclusion}
In this paper, a comprehensive review of the application of GNNs for traffic forecasting is presented. Three levels of traffic problems and graphs are summarized, namely, road-level, region-level and station-level. The usages of recurrent GNNs, convolutional GNNs and graph autoencoders are discussed. We also give the latest collection of open dataset and code resource for this topic. Challenges and future directions are further pointed out for the follow-up research.

\bibliography{references}

\end{document}